\documentclass{article}

\usepackage{amsmath,amssymb}
\usepackage[mathscr]{euscript} 
\usepackage{amsthm} 
\usepackage{tikz} 
\usepackage{graphicx} 
\newcommand*{\Scale}[2][4]{\scalebox{#1}{\ensuremath{#2}}} 
\usepackage{framed} 
\usepackage{float} 
\usepackage{multirow}
\usepackage{color}
\usepackage{subfigure}
\usepackage{longtable}
\usepackage{cite}
\usepackage[hidelinks, breaklinks=true]{hyperref} 
\usepackage{xcolor}  \definecolor{shadecolor}{rgb}{.95,.95,.95}  
\usepackage[font=footnotesize]{caption}
\usepackage{nicefrac}
\usepackage{dsfont} 
\usepackage[export]{adjustbox} 
\usepackage{cancel} 
\usepackage{enumitem}	

\usepackage[ruled,vlined]{algorithm2e}
\usepackage{algorithmic}

\newtheorem{myTheorem}{Theorem}
\newtheorem{myLemma}{Lemma}

\newtheorem{myExample}{Example}

\newtheorem{myRemark}{Remark}

\tikzstyle{every edge}=  [draw]
\tikzstyle{vertex} = [draw,circle,minimum size=1pt]
\tikzstyle{label} = [minimum size=.1pt,font=\scriptsize]
\tikzstyle{title} = [minimum size=.25cm,font=\small]

\newcommand{\bs}[1]{\boldsymbol{#1}}
\newcommand{\bhat}[1]{\boldsymbol{\hat{#1}}}
\newcommand{\btilde}[1]{\boldsymbol{\tilde{#1}}}

\newcommand{\magenta}{\textcolor{magenta}}
\newcommand{\cyan}{\textcolor{cyan}}
\newcommand{\gray}{\textcolor{gray}}
\newcommand{\white}{\textcolor{white}}

\def \R{\mathbb{R}}
\def \Pr{\mathbb{P}}

\def \E{\mathbb{E}}

\def \1{{\mathds{1}}}
\def \T{\mathsf{T}}

\def \spn{{\rm span}}
\def \Ord{\mathscr{O}}
\def \<{\langle}
\def \>{\rangle}

\DeclareMathOperator*{\argmax}{arg\,max}

\def \d{{\hyperref[dDef]{{\rm d}}}}
\def \r{{\hyperref[rDef]{{\rm r}}}}
\def \Kk{{\hyperref[KkDef]{{\rm K}}}}
\def \n{{\hyperref[nDef]{{\rm n}}}}
\def \m{{\hyperref[mDef]{{\rm m}}}}
\def \L{{\hyperref[LDef]{\ell}}}
\def \xi{{\hyperref[xiDef]{{\rm x}}}}
\def \p{{\hyperref[pDef]{{\rm p}}}}
\def \deltaa{{\hyperref[deltaaDef]{\delta}}}

\def \x{{\hyperref[xDef]{\bs{{\rm x}}}}}
\def \btheta{{\hyperref[bthetaDef]{\bs{\theta}}}}

\def \X{{\hyperref[XDef]{\bs{{\rm X}}}}}
\def \Z{{\hyperref[ZDef]{\bs{{\rm Z}}}}}
\def \U{{\hyperref[UDef]{\bs{{\rm U}}}}}
\def \bTheta{{\hyperref[bThetaDef]{\bs{\Theta}}}}
\def \P{{\hyperref[PDef]{\bs{{\rm P}}}}}

\def \i{{\hyperref[iDef]{{\rm i}}}}
\def \j{{\hyperref[jDef]{{\rm j}}}}
\def \k{{\hyperref[kDef]{{\rm k}}}}
\def \tauu{{\hyperref[tauuDef]{\tau}}}

\def \O{{\hyperref[ODef]{\bs{\Omega}}}}
\def \o{{\hyperref[oDef]{\bs{\omega}}}}
\def \ups{{\hyperref[upsDef]{\bs{\upsilon}}}}

\def \Aone{{\hyperref[AoneDef]{{\rm {\bf A1}}}}}
\def \Atwo{{\hyperref[AtwoDef]{{\rm {\bf A2}}}}}

\def \PCA{{\hyperref[PCADef]{PCA}}}
\def \HRMC{{\hyperref[HRMCDef]{HRMC}}}
\def \LRMC{{\hyperref[LRMCDef]{LRMC}}}
\def \MMC{{\hyperref[MMCDef]{MMC}}}
\def \AMMC{{\hyperref[AMMCDef]{AMMC}}}
\def \SC{{\hyperref[SCDef]{SC}}}

\newcommand*{\titleGP}{\begingroup 
\centering 
\vspace*{\baselineskip} 

\rule{\textwidth}{1.6pt}\vspace*{-\baselineskip}\vspace*{2pt} 
\rule{\textwidth}{0.4pt}\\[.5\baselineskip] 

{\LARGE \vspace{.3cm} Mixture Matrix Completion} \\ [0.4\baselineskip] 

\rule{\textwidth}{0.4pt}\vspace*{-\baselineskip}\vspace{3.2pt} 
\rule{\textwidth}{1.6pt}\\[\baselineskip] 

{\scshape 
Daniel L. Pimentel-Alarc\'on}

{\itshape Georgia State University \par}
\endgroup}

\begin{document}
\titleGP

\begin{abstract}
Completing a data matrix $\X$ has become an ubiquitous problem in modern data science, with applications in recommender systems, computer vision, and networks inference, to name a few. One typical assumption is that $\X$ is low-rank. A more general model assumes that each {\em column} of $\X$ corresponds to one of several low-rank matrices. This paper generalizes these models to what we call {\em mixture matrix completion} (\MMC): the case where each {\em entry} of $\X$ corresponds to one of several low-rank matrices. \MMC\ is a more accurate model for recommender systems, and brings more flexibility to other completion and clustering problems. We make four fundamental contributions about this new model. First, we show that \MMC\ is theoretically possible (well-posed). Second, we give its precise information-theoretic identifiability conditions. Third, we derive the sample complexity of \MMC. Finally, we give a practical algorithm for \MMC\ with performance comparable to the state-of-the-art for simpler related problems, both on synthetic and real data.
\end{abstract}

\section{Introduction}
Matrix completion aims to estimate the missing entries of an incomplete data matrix $\X$. One of its main motivations arises in recommender systems, where each row represents an item, and each column represents a user. We only observe an entry in $\X$ whenever a user rates an item, and the goal is to predict unseen ratings in order to make good recommendations.

{\bf Related Work.} In 2009, Cand\`es and Recht \cite{candes-recht} introduced \phantomsection\label{LRMCDef}{\em low-rank matrix completion} (\LRMC), arguably the most popular model for this task. \LRMC\ assumes that each column (user) can be represented as a linear combination of a few others, whence $\X$ is low-rank. Later in 2012, Eriksson et.~al.~\cite{HRMC} introduced \phantomsection\label{HRMCDef}{\em high-rank matrix completion} (\HRMC), also known as {\em subspace clustering with missing data}. This more general model assumes that each column of $\X$ comes from one of several low-rank matrices, thus allowing several types of users. Since their inceptions, both \LRMC\ and \HRMC\ have attracted a tremendous amount of attention (see \cite{candes-recht, candes-tao, svt, keshavan10, grouse, recht, fpc, lmafit, altLRMC, lmafit2, coherent, iterative, incoherent, kiraly, LRMCpimentel, identifiability, converse, balzano, HRMC, ssp14, yang, elhamifarNIPS, ongie, gssc, infoTheoretic,aggarwal1,aggarwal2} for a very incomplete list).

\begin{figure}
\centering
\includegraphics[width=\textwidth]{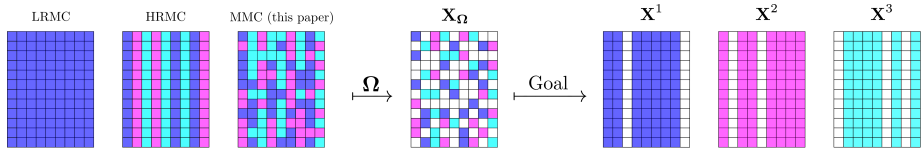}
\caption{In \LRMC, $\X$ is a low-rank matrix. In \HRMC, each {\em column} of $\X$ comes from one of several low-rank matrices. In \MMC, each {\em entry} comes from one of several low-rank matrices $\X{}^1,\dots,\X{}^\Kk$; we only observe $\X_\O$, and our goal is to recover the columns of $\X{}^1,\dots,\X{}^\Kk$ that have observations in $\X_\O$.}
\label{generalizationMMCFig}
\end{figure}

{\bf Paper contributions.} This paper introduces an even more general model: \phantomsection\label{MMCDef}{\em mixture matrix completion} (\MMC), which assumes that each {\em entry} in $\X$ (rather than column) comes from one out of several low-rank matrices, and the goal is to recover the matrices in the mixture. Figure \ref{generalizationMMCFig} illustrates the generalization from \LRMC\ to \HRMC\ and to \MMC.  One of the main motivations behind \MMC\ is that users often share the same account, and so each column in $\X$ may contain ratings from several users. Nonetheless, as we show in Section \ref{applicationsSec}, \MMC\ is also a more accurate model for many other contemporary applications, including networks inference, computer vision, and metagenomics. This paper makes several fundamental contributions about \MMC:

\begin{itemize}[leftmargin=0.4cm]
\item[--]
{\bf Well posedness.} First, we show that \MMC\ is theoretically possible if we observe {\em the right entries} and the mixture is {\em generic} (precise definitions below).
\item[--]
{\bf Identifiability conditions.} We provide precise information-theoretical conditions on the entries that need to be observed such that a mixture of $\Kk$ low-rank matrices is identifiable. These extend similar recent results of \LRMC\ \cite{LRMCpimentel} and \HRMC\ \cite{infoTheoretic} to the setting of \MMC. The subtlety in proving these results is that there could exist {\em false} mixtures that agree with the observed entries, even if the sampling is uniquely completable for \LRMC\ and \HRMC\ (see Example \ref{identifiabilityEg}). In other words, there exits samplings that are identifiable for \LRMC\ (and \HRMC) but are not identifiable for \MMC, and so in general it is not enough to simply have $\Kk$ times more samples. Hence, it was necessary to derive identifiability conditions for \MMC, similar to those of \LRMC\ in \cite{LRMCpimentel} and \HRMC\ in \cite{infoTheoretic}. We point out that in contrast to typical completion theory \cite{candes-recht, HRMC, candes-tao, svt, grouse, keshavan10, recht, fpc, lmafit, altLRMC, lmafit2, coherent, iterative, incoherent, balzano, ssp14, gssc, yang}, these type of identifiability conditions are deterministic (not restricted to uniform sampling), and make no coherence assumptions.
\item[--]
{\bf Sample complexity.} If $\X \in \R{}^{\d \times \n}$ is a mixture of $\Kk$ rank-$\r$ matrices, we show that with high probability, our identifiability conditions will be met if each entry is observed with probability $\Ord(\frac{\Kk}{\d}\max\{\r,\log\d\})$, thus deriving the sample complexity of \MMC, which is the same as the sample complexity of \HRMC\ \cite{infoTheoretic}, and simplifies to $\Ord(\frac{1}{\d}\max\{\r,\log\d\})$ in the case of $\Kk=1$, which corresponds to the sample complexity of \LRMC\ \cite{LRMCpimentel}. Intuitively, this means that information-theoretically, we virtually pay no price for mixing low-rank matrices.
\item[--]
{\bf Practical algorithm.} Our identifiability results follow from a combinatorial analysis that is infeasible in practice. To address this, we give a practical alternating algorithm for \MMC\ whose performance (in the more difficult problem of \MMC) is comparable to state-of-the-art algorithms for the much simpler problems of \HRMC\ and \LRMC.
\end{itemize}

\section{Motivating Applications}
\label{applicationsSec}
Besides recommender systems, there are many important applications where data can be modeled as a mixture of low-rank matrices. Here are a few examples motivated by current data science challenges.

{\bf Networks Inference.}
Estimating the topology of a network (internet, sensor networks, biological networks, social networks) has been the subject of a large body of research in recent years \cite{network, network2, network3, network4, network5, network6, network7}. To this end, companies routinely collect distances between nodes (e.g., computers) that connect with monitors (e.g., Google, Amazon, Facebook) in a data matrix $\X$.  In a simplified model, if node $\j$ is in subnet $\k$, then the $\j{}^{\rm th}$ column can be modeled as the sum of (i) the distance between node $\j$ and router $\k$, and (ii) the distance between router $\k$ and each of the monitors. Hence, the columns (nodes) corresponding to each subnet form a low-rank matrix, which is precisely the model assumed by \HRMC. However, depending on the network's traffic, each node may use different routes to communicate at different times. Consequently, the same column in $\X$ may contain measurements from different low-rank matrices. In other words, distance matrices of networks are a mixture of low-rank matrices.

{\bf Computer Vision.}
Background segmentation is one of the most fundamental and crucial tasks in computer vision, yet it can be tremendously challenging. The vectorized frames of a video can be modeled as columns with some entries (pixels) in a low-rank background, and some outlier entries, corresponding to the foreground. Typical methods, like the acclaimed \phantomsection\label{PCADef}robust \PCA\ (principal component analysis) \cite{robustpca, review, r2pca, alm, almNIPS, rpcaWebsite, brpca, rosl, iht, survey, alternating, apg}, assume that the foreground is sparse and has no particular structure. However, in many situations this is not the case.  For instance, since the location of an object in consecutive frames is highly correlated, the foreground can be highly structured. Similarly, the foreground may not be sparse, specially if there are foreground objects moving close to the camera (e.g., in a selfie). Even state-of-the-art methods fail in scenarios like these, which are not covered by current models (see Figure \ref{yaleFig} for an example). In contrast, \MMC\ allows to use one matrix in the mixture to represent the background, other matrices to represent foreground objects (small or large, even dominant), and even other matrices to account for occlusions and other illumination/visual artifacts. Hence, \MMC\ can be a more accurate model for video segmentation and other image processing tasks, including inpainting \cite{inpainting} and face clustering, which we explore in our experiments.

{\bf Metagenomics.}
One contemporary challenge in Biology is to quantify the presence of different types of bacteria in a system (e.g., the human gut microbiome) \cite{metagenomics1, metagenomics2, metagenomics3, metagenomics4, metagenomics5}. The main idea is to collect several DNA samples from such a system, and use their genomic information to count the number of bacteria of each type (the genome of each bacterium determines its type).  In practice, to obtain an organism's genome (e.g., a person's genome), biologists feed a DNA sample (e.g., blood or hair) to a sequencer machine that produces a series of {\em reads}, which are short genomic sequences that can later be assembled and aligned to recover the entire genome. The challenge arises when the sequencer is provided a sample with DNA from multiple organisms, as is the case in the human gut microbiome, where any sample will contain a mixture of DNA from multiple bacteria that cannot be disentangled into individual bacterium. In this case, each read produced by the sequencer may correspond to a different type of bacteria. Consequently, each DNA sample (column) may contain genes (rows) from different types of bacteria, which is precisely the model that \MMC\ describes.

\section{Problem Statement}
\label{problemSec}
Let \phantomsection\label{XDef}\phantomsection\label{dDef}\phantomsection\label{nDef}\phantomsection\label{KkDef}$\X{}^1,\dots,\X{}^\Kk \in \R{}^{\d \times \n}$ be a set of rank-\phantomsection\label{rDer}$\r$ matrices, and let \phantomsection\label{ODef}$\O{}^1,\dots,\O{}^\k \in \{0,1\}{}^{\d \times \n}$ indicate {\em disjoint} sets of observed entries. Suppose $\X{}^1,\dots,\X{}^\Kk$ and $\O{}^1,\dots,\O{}^\Kk$ are unknown, and we only observe $\X_\O$, defined as follows:
\begin{itemize}
\item[--]
If the $(\i,\j){}^{\rm th}$ entry of $\O{}^\k$ is $1$, then the $(\i,\j){}^{\rm th}$ entry of $\X_\O$ is equal to the $(\i,\j){}^{\rm th}$ entry of $\X{}^\k$.
\item[--]
If the $(\i,\j){}^{\rm th}$ entry of $\O{}^\k$ is $0$ for every \phantomsection\label{kDef}$\k=1,\dots,\Kk$, then the $(\i,\j){}^{\rm th}$ entry of $\X_\O$ is missing.
\end{itemize}
This way $\O{}^\k$ indicates the entries of $\X_\O$ that correspond to $\X{}^\k$, and $\O:=\sum_{\k=1}^\Kk \O{}^\k$ indicates the set of {\em all} observed entries. Since $\O{}^1,\dots,\O{}^\Kk$ are disjoint, $\O \in \{0,1\}{}^{\d \times \n}$. Equivalently, each observed entry of $\X_\O$ corresponds to an entry in either $\X{}^1$ or $\X{}^2$ or $\dots$ or $\X{}^\Kk$ (i.e., there are no {\em collisions}).  In words, $\X_\O$ contains a {\em mixture} of entries from several low-rank matrices.

The goal of \MMC\ is to recover all the columns of $\X{}^1,\dots,\X{}^\Kk$ that have observations in $\X_\O$ (see Figure \ref{generalizationMMCFig} to build some intuition). In our recommendations example, a column \phantomsection\label{xDef}$\x_\o \in \X_\O$ will contain entries from $\X{}^\k$ whenever $\x_\o$ contains ratings from a user of the $\k{}^{\rm th}$ type. Similarly, the same column will contain entries from \phantomsection\label{LDef}$\X{}^\L$ whenever it also contains ratings from a user of the $\L{}^{\rm th}$ type. We would like to predict the preferences of both users, or more generally, all users that have ratings in $\x_\o$.  On the other hand, if $\x_\o$ has no entries from $\X{}^\k$, then $\x_\o$ involves no users of the $\k{}^{\rm th}$ type, and so it would be impossible (and futile) to try to recover such column of $\X{}^\k$. In \MMC, the matrices $\O{}^1,\dots,\O{}^\Kk$ play the role of the {\em hidden} variables constantly present in mixture problems. Notice that if we knew $\O{}^1,\dots,\O{}^\Kk$, then we could partition $\X_\O$ accordingly, and estimate $\X{}^1,\dots,\X{}^\Kk$ using standard \LRMC. The challenge is that we do not know $\O{}^1,\dots,\O{}^\Kk$.

\subsection{The Subtleties of MMC}
\label{subtletiesSec}
The main theoretical difficulty of \MMC\ is that depending on the pattern of missing data, there could exist {\em false} mixtures. That is, matrices $\btilde{\X}{}^1,\dots,\btilde{\X}{}^\Kk$, other than $\X{}^1,\dots,\X{}^\Kk$, that agree with $\X_\O$, even if $\X{}^1,\dots,\X{}^\Kk$ are observed on uniquely completable patterns for \LRMC.

\begin{myExample}
\label{identifiabilityEg}
Consider the next rank-$1$ matrices $\X{}^1,\X{}^2$, and their partially observed mixture $\X_\O$:
\begin{align*}
\X{}^1=\left[ \cyan{\begin{matrix}
1 & 2 & 3 & 4 \\
1 & 2 & 3 & 4 \\
1 & 2 & 3 & 4 \\
1 & 2 & 3 & 4 \\
1 & 2 & 3 & 4
\end{matrix}} \right],
\hspace{.5cm}
\X{}^2=\left[ \magenta{ \begin{matrix}
1 & 2 & 3 & 4 \\
2 & 4 & 6 & 8 \\
3 & 6 & 9 & 12 \\
4 & 8 & 12 & 16 \\
5 & 10 & 15 & 20
\end{matrix}} \right],
\hspace{.5cm}
\X_\O \ = \ \left[ \begin{matrix}
\cyan{1} & \cdot & \magenta{3} & \magenta{4} \\
\cyan{1} & \cyan{2} & \cdot & \magenta{8} \\
\magenta{3} & \cyan{2} & \cyan{3} & \cdot \\
\magenta{4} & \magenta{8} & \cyan{3} & \cyan{4} \\
\cdot & \magenta{10} & \magenta{15} & \cyan{4}
\end{matrix} \right].
\end{align*}
We can verify that $\X{}^1$ and $\X{}^2$ are observed on uniquely completable sampling patterns for \LRMC\ \cite{LRMCpimentel}. Nonetheless, we can construct the following {\em false} rank-$1$ matrices that agree with $\X_\O$:
\small
\begin{align*}
\btilde{\X}{}^1 = \left[ \gray{\begin{matrix}
60 & 40 & 15 & \magenta{4} \\
\cyan{1} & \nicefrac{2}{3} & \nicefrac{1}{4} & \nicefrac{1}{15} \\
\magenta{3} & \cyan{2} & \nicefrac{3}{4} & \nicefrac{1}{5} \\
12 & \magenta{8} & \cyan{3} & \nicefrac{4}{5} \\
60 & 40 & \magenta{15} & \cyan{4}
\end{matrix}} \right],
\hspace{.2cm}
\btilde{\X}{}^2 = \left[ \gray{ \begin{matrix}
\cyan{1} & \nicefrac{1}{4} & \magenta{3} & 1 \\
8 & \cyan{2} & 24 & \magenta{8} \\
1 & \nicefrac{1}{4} & \cyan{3} & 1 \\
\magenta{4} & 1 & 12 & \cyan{4} \\
40 & \magenta{10} & 120 & 40
\end{matrix}} \right].
\end{align*}
\normalsize
This shows that even with unlimited computational power, if we exhaustively search all the identifiable patterns for \LRMC, we can end up with false mixtures. Hence the importance of studying the identifiable patterns for \MMC.
\end{myExample}

False mixtures arise because we do not know a priori which entries of $\X_\O$ correspond to each $\X{}^\k$. Hence, it is possible that a rank-$\r$ matrix $\btilde{\X}$ agrees with some entries from $\X{}^1$, other entries from $\X{}^2$, and so on. Furthermore, $\btilde{\X}$ may even be {\em the only} rank-$\r$ matrix that agrees with such combination of entries, as in Example \ref{identifiabilityEg}.

\begin{myRemark}
\label{difficultyRmk}
Recall that \LRMC\ and \HRMC\ are tantamount to identifying the subspace(s) containing the columns of $\X$ \cite{LRMCpimentel,infoTheoretic}. In fact, if we knew such subspaces, \LRMC\ and \HRMC\ become almost trivial problems (see Appendix \ref{furtherApx} for details). Similarly, if no data is missing, \HRMC\ simplifies to subspace clustering, which has been studied extensively, and is now reasonably well-understood \cite{sc,liu1, liu2, mahdi, qu, peng,wang,aarti,hu,ssc}. In contrast, \MMC\ remains challenging even if the subspaces corresponding to the low-rank matrices in the mixture are known, and even $\X$ is fully observed. We refer the curious reader to Appendix \ref{furtherApx}, and point out the bottom row and the last column in Figure \ref{ammcFig}, which show the \MMC\ error when the underlying subspaces are known, and when $\X$ is fully observed.
\end{myRemark}

\section{Main Theoretical Results}
\label{mainResultSec}
Example \ref{identifiabilityEg} shows the importance of studying the identifiable patterns for \MMC, which we do now. First recall that $\r+1$ samples per column are necessary for \LRMC\ \cite{LRMCpimentel}. This implies that even if an oracle told us $\O{}^1,\dots,\O{}^\Kk$, if we intend to recover a column of $\X{}^\k$, we need to observe it on at least $\r+1$ entries. Hence we assume without loss of generality that:
\begin{shaded*}
\begin{itemize}
\phantomsection\label{AoneDef}
\item[(\Aone)]
Each column of $\O{}^\k$ has either $0$ or $\r+1$ non-zero entries.
\end{itemize}
\end{shaded*}
In words, \Aone\ requires that each column of $\X{}^\k$ to be recovered is observed on exactly $\r+1$ entries. Of course, observing more entries may only aid completion. Hence, rather than an assumption, \Aone\ describes the most difficult scenario where we have the bare minimum amount of information required for completion. We use \Aone\ to ease notation, exposition and analysis. All our results can be easily extended to the case where \Aone\ is droped (see Remark \ref{AoneRmk}).

Without further assumptions on $\X$, completion (of any kind) may be impossible.  To see this consider the simple example where $\X$ is only supported on the \phantomsection\label{iDef}$\i{}^{\rm th}$ row. Then it would be impossible to recover $\X$ unless all columns were observed on the $\i{}^{\rm th}$ row. In most completion applications this would be unlikely. For example, in a movies recommender system like Netflix, this would require that {\em all} the users watched (and rated) the same movie.

To rule out scenarios like these, typical completion theory requires incoherence and uniform sampling. Incoherence guarantees that the information is well-spread over the matrix. Uniform sampling guarantees that all rows and columns are sufficiently sampled. However, it is usually unclear (and generally unverifiable) whether an incomplete matrix is coherent.  Furthermore, observations are hardly ever uniformly distributed. For instance, we do not expect children to watch adults movies.

To avoid these issues, instead of incoherence we will assume that $\X$ is a {\em generic} mixture of low-rank matrices. More precisely, we assume that:
\begin{shaded*}
\begin{itemize}
\phantomsection\label{AtwoDef}
\item[(\Atwo)]
$\X{}^1,\dots,\X{}^\Kk$ are drawn independently according to an absolutely continuous distribution with respect to the Lebesgue measure on the determinantal variety (set of all $\d \times \n$, rank-$\r$ matrices).
\end{itemize}
\end{shaded*}
\Atwo\ essentially requires that each $\X{}^\k$ is a generic rank-$\r$ matrix. This type of {\em genericity} assumptions are becoming increasingly common in studies of \LRMC, \HRMC, and related problems \cite{LRMCpimentel, infoTheoretic, identifiability, kiraly, converse, aggarwal1, aggarwal2, r2pca}. See Appendix \ref{assumptionsSec} for a further discussion on \Atwo, and its relation to other common assumptions from the literature.

With this, we are ready to present our main theorem. It gives a deterministic condition on $\O$ to guarantee that $\X{}^1,\dots,\X{}^\Kk$ can be identified from $\X_\O$. This provides information-theoretic requirements for \MMC. The proof is in Appendix \ref{proofApx}.

\begin{framed}
\begin{myTheorem}
\label{mainThm}
Let \Aone-\Atwo\ hold. Suppose there exist matrices \phantomsection\label{tauuDef}$\{\O_\tauu\}_{\tauu=1}^{\r+1}$ formed with disjoint subsets of $(\d-\r+1)$ columns of $\O{}^\k$, such that for every $\tauu$:
\begin{itemize}
\item[$(\dagger)$]
Every matrix $\O'$ formed with a {\em proper} subset of the columns in $\O_\tauu$ has at least $\r$ fewer columns than non-zero rows.
\end{itemize}
Then all the columns of $\X{}^\k$ that have observations in $\X_\O$ are identifiable.
\end{myTheorem}
\end{framed}

In words, Theorem \ref{mainThm} states that \MMC\ is possible as long as we observe {\em the right entries} in each $\X{}^\k$. The intuition is that each of these entries imposes a constraint on what $\X{}^1,\dots,\X{}^\Kk$ may be, and the pattern in $\O$ determines whether these constraints are redundant. Patterns satisfying the conditions of Theorem \ref{mainThm} guarantee that $\X{}^1,\dots,\X{}^\Kk$ is the only mixture that satisfies the constraints produced by the observed entries.

\begin{myRemark}
\label{AoneRmk}
Recall that $\r+1$ samples per column are strictly necessary for completion. \Aone\ requires that we have exactly that minimum number of samples. If $\X{}^\k$ is observed on more than $\r+1$ entries per column, it suffices that $\O{}^\k$ contains a pattern satisfying the conditions of Theorem \ref{mainThm}.
\end{myRemark}

Theorem \ref{mainThm} shows that \MMC\ is possible if the samplings satisfy certain combinatorial conditions. Our next result shows that if each entry of $\X{}^\k$ is observed on $\X_\O$ with probability $\Ord(\frac{1}{\d}\max\{\r,\log\d\})$, then with high probability $\O{}^\k$ will satisfy such conditions. The proof is in Appendix \ref{proofApx}.

\begin{framed}
\begin{myTheorem}
\label{probabilityThm}
Suppose $\r \leq \frac{\d}{6}$ and $\n \geq (\r+1)(\d-\r+1)$.  Let $\epsilon >0$ be given. Suppose that an entry of $\X_\O$ is equal to the corresponding entry of $\X{}^\k$ with probability\phantomsection\label{pDef}
\begin{align*}
\textstyle \p \ \geq \ \frac{2}{\d} \max \left\{2\r, \ 12\left( \log(\frac{\d}{\epsilon})+1\right) \right\}.
\end{align*}
Then $\O{}^\k$ satisfies the sampling conditions of Theorem \ref{mainThm} with probability $\geq 1-2(\r+1)\epsilon$.
\end{myTheorem}
\end{framed}

Theorem \ref{probabilityThm} shows that the sample complexity of \MMC\ is $\Ord(\Kk\max\{\r,\log\d\})$ observations per column of $\X_\O$. This is exactly the same as the sample complexity of \HRMC\ \cite{infoTheoretic}, and simplifies to $\Ord(\max\{\r,\log\d\})$ if $\Kk=1$, corresponding to the sample complexity of \LRMC\ \cite{LRMCpimentel}. Intuitively, this means that information-theoretically, we virtually pay no price for mixing low-rank matrices.

\section{Alternating Algorithm for MMC}
\label{algorithmSec}
Theorems \ref{mainThm} and \ref{probabilityThm} show that \MMC\ is theoretically possible under reasonable conditions (virtually the same as \LRMC\ and \HRMC).  However, these results follow from a combinatorial analysis that is infeasible in practice (see Appendix \ref{proofApx} for details). To address this, we derive a practical alternating algorithm for \MMC, which we call \phantomsection\label{AMMCDef}AMMC (alternating mixture matrix completion).

The main idea is that \MMC, like most mixture problems, can be viewed as a clustering task: if we could determine the entries of $\X_\O$ that correspond to each $\X{}^\k$, then we would be able to partition $\X_\O$ into $\Kk$ incomplete low-rank matrices, and then complete them using standard \LRMC. The question is how to determine which entries of $\X_\O$ correspond to each $\X{}^\k$, i.e., how to determine $\O{}^1,\dots,\O{}^\Kk$.
To address this, let \phantomsection\label{UDef}$\U{}^\k \in \R{}^{\d \times \r}$ be a basis for the subspace containing the columns of $\X{}^\k$, and let $\x_\o$ denote the \phantomsection\label{jDef}$\j{}^{\rm th}$ column of $\X_\O$, observed only on the entries indexed by \phantomsection\label{oDef}$\o \subset \{1,\dots,\d\}$. For any subspace, matrix or vector that is compatible with a set of indices $\bs{\cdot}$, we use the subscript $\bs{\cdot}$ to denote its restriction to the coordinates/rows in $\bs{\cdot}$.  For example, $\U{}^\k_\o \in \R{}^{|\o| \times \r}$ denotes the restriction of $\U{}^\k$ to the indices in $\o$. Suppose $\x_\o$ contains entries from $\X{}^\k$, and let $\o{}^\k \subset \o$ index such entries. Then our goal is to determine $\o{}^\k$, as that would tell us the $\j{}^{\rm th}$ column of $\O{}^\k$. Since $\x_{\o{}^\k} \in \spn\{\U{}^\k_{\o{}^\k}\}$, we can restate our goal as finding the set $\o{}^\k \subset \o$ such that $\x_{\o{}^\k} \in \spn\{\U{}^\k_{\o{}^\k}\}$.

To find $\o{}^\k$, let \phantomsection\label{upsDef}$\ups \subset \o$, and let \phantomsection\label{PDef}$\P{}^\k_\ups:=\U{}^\k_\ups(\U{}^{\k\T}_\ups\U{}^\k_\ups){}^{-1}\U{}^{\k\T}_\ups$ denote the projection operator onto $\spn\{\U{}^\k_\ups\}$. Recall that $\|\P{}^\k_\ups \x_\ups\| \leq \|\x_\ups\|$, with equality if and only if $\x_\ups \in \spn\{\U{}^\k_\ups\}$. It follows that $\o{}^\k$ is the largest set $\ups$ such that $\|\P{}^\k_\ups \x_\ups\| = \|\x_\ups\|$. In other words, $\o{}^\k$ is the solution to
\begin{align}
\label{idealEq}
\argmax_{\ups \subset \o} \ \ \|\P{}^\k_\ups \x_\ups\| - \|\x_\ups\| \ + \ |\ups|.
\end{align}
However, \eqref{idealEq} is non-convex. Hence, in order to find the solution to \eqref{idealEq}, we propose the following {\em erasure} strategy. The main idea is to start our search with $\ups=\o$, and then iteratively remove the entries (coordinates) of $\ups$ that most increase the gap between $\|\P{}^\k_\ups \x_\ups\|$ and $\|\x_\ups\|$ (hence the term {\em erasure}). We stop this procedure when $\|\P{}^\k_\ups \x_\ups\|$ is equal to $\|\x_\ups\|$ (or close enough). More precisely, we initialize $\ups=\o$, and then iteratively redefine $\ups$ as the set
\begin{align}
\label{erasureEq}
\ups \ = \ \ups \backslash \i,
\hspace{.5cm} \text{where} \hspace{.5cm}
\i \ = \ \argmax_{i \in \ups } \ \ \| \P{}^\k_{\ups \backslash i} \x_{\ups \backslash i} \| - \|\x_{\ups \backslash i}\|.
\end{align}
In words, $\i$ is the coordinate of the vector $\x_\ups$ such that if ignored, the gap between the remaining vector $\x_{\ups \backslash \i}$ and its projection $\P{}^\k_{\ups \backslash \i} \x_{\ups \backslash \i}$ is reduced the most. At each iteration we remove (erase) such coordinate $\i$ from $\ups$. The intuition behind this approach is that the coordinates of $\x_\ups$ that do not correspond to $\X{}^\k$ are more likely to increase the gap between $\|\P{}^\k_\ups \x_\ups\|$ and $\|\x_\ups\|$.
Notice that if $\U{}^\k$ is in general position (guaranteed by \Atwo) and $|\ups| \leq \r$, then $\U{}^\k_\ups = \R{}^{|\ups|}$ (because $\U{}^\k$ is $\r$-dimensional). In such case, it is trivially true that $\x_\ups \in \spn\{\U{}^\k_\ups\}$, whence $\|\P{}^\k_\ups \x_\ups\| = \|\x_\ups\|$. Hence the procedure above is guaranteed to terminate after at most $\|\o\|-\r$ iterations. At such point, $|\ups|=\r$, and we know that we were unable to find $\o{}^\k$ (or a subset of it). One alternative is to start with a different $\ups_0 \subsetneq \o$, and search again.

This procedure may remove some entries from $\o{}^\k$ along the way, so in general, the output of this process will be a set $\ups \subset \o{}^\k$. However, finding a subset of $\o{}^\k$ is enough to find $\o{}^\k$. To see this, recall that since $\x_{\o{}^\k} \in \spn\{\U{}^\k_{\o{}^\k}\}$, there is a coefficient vector \phantomsection\label{bthetaDef}$\btheta{}^\k \in \R{}^\r$ such that $\x_{\o{}^\k}=\U{}^\k_{\o{}^\k} \btheta{}^\k$. Since $\ups \subset \o{}^\k$, it follows that $\x_\ups=\U{}^\k_\ups\btheta{}^\k$. Furthermore, since $|\ups| \geq \r$, we can find $\btheta{}^\k$ as $\btheta{}^\k = (\U{}^{\k\T}_\ups \U{}^\k_\ups){}^{-1}\U{}^{\k\T}_\ups \x_\ups$.  Since $\x_{\o{}^\k}=\U{}^\k_{\o{}^\k} \btheta{}^\k$, at this point we can identify $\o{}^\k$ by simple inspection (the matching entries in $\x_\o$ and $\U{}^\k_\o \btheta{}^\k$).
%
Recall that $\o{}^\k$ determines the $\j{}^{\rm th}$ column of $\O{}^\k$. Hence, if we repeat the procedure above for each column in $\X_\O$ and each $\k$, we can recover $\O{}^1,\dots,\O{}^\Kk$. After this, we can use standard \LRMC\ on $\X_{\O{}^1},\dots,\X_{\O{}^\Kk}$ to recover $\X{}^1,\dots\X{}^\Kk$ (which is the ultimate goal of \MMC).

The catch here is that this procedure requires knowing $\U{}^\k$, which we do not know. So essentially we have a chicken and egg problem: (i) if we knew $\U{}^\k$, we would be able to find $\O{}^\k$. (ii) If we knew $\O{}^\k$ we would be able to find $\U{}^\k$ (and $\X{}^\k$, using standard \LRMC\ on $\X_{\O{}^\k}$). Since we know neither, we use a common technique for these kind of problems: alternate between finding $\O{}^\k$ and $\U{}^\k$.
More precisely, we start with some initial guesses $\bhat{\U}{}^1,\dots,\bhat{\U}{}^\Kk$, and then alternate between the following two steps until convergence:
\begin{itemize}[leftmargin=0.6cm]
\item[(i)]
\textbf{Cluster.}
Let $\x_\o$ be the $\j{}^{\rm th}$ column in $\X_\O$. For each $\k=1,\dots,\Kk$, we first {\em erase} entries from $\o$ to obtain a set $\ups \subset \o$ indicating entries likely to correspond to $\X{}^\k$. This {\em erasure} procedure initializes $\ups=\o$, and then repeats \eqref{erasureEq}, (replacing $\P{}^\k$ with $\bhat{\P}{}^\k$, which denotes the projection operator onto $\spn\{\bhat{\U}{}^\k\}$) until we to obtain a set $\ups \subset \o$ such that the projection $\|\bhat{\P}{}^\k_{\ups} \x_{\ups}\|$ is close to $\|\x_{\ups}\|$. This way, the entries of $\x_{\ups}$ are likely to correspond to $\X{}^\k$.
Using these entries, we can estimate the coefficient of the $\j{}^{\rm th}$ column of $\X{}^\k$ with respect to $\U{}^\k$, given by $\bhat{\btheta}{}^\k = (\bhat{\U}{}^{\k\T}_{\ups{}^\k} \bhat{\U}{}^\k_{\ups{}^\k}){}^{-1} \bhat{\U}{}^{\k\T}_{\ups{}^\k} \x_{\ups{}^\k}$. With $\bhat{\btheta}{}^\k$ we can also estimate the $\j{}^{\rm th}$ column of $\X{}^\k$ as $\bhat{\x}{}^\k := \bhat{\U}{}^\k \bhat{\btheta}{}^\k$.
%
%
Notice that both $\ups$ and $\bhat{\x}{}^\k$ are obtained using $\bhat{\U}{}^\k$, which may be different from $\U{}^\k$. It follows that $\ups$ may contain some entries that do not correspond to $\X{}^\k$, and $\bhat{\x}{}^\k$ may be inaccurate. Hence, in general, $\x_\o$ and $\bhat{\x}{}^\k_\o$ will have no matching entries, and so we cannot identify $\o{}^\k$ by simple inspection, as before.
However, we can repeat our procedure for each $\k$ to obtain estimates $\bhat{\x}{}^1_\o,\dots,\bhat{\x}{}^\Kk_\o$, and then assign each entry of $\x_\o$ to its closest match. More precisely, our estimate $\bhat{\o}{}^\k \subset \o$ (indicating the entries of $\x_\o$ that we estimate that correspond to $\X{}^\k$) will contain entry $\i \in \o$ if $|{\rm x}_\i - \hat{{\rm x}}{}^\k_\i| \leq |{\rm x}_\i - \hat{{\rm x}}{}^{\L}_\i|$ for every $\L=1,\dots,\Kk$.
%
%
Repeating this procedure for each column of $\X_\O$ will produce estimates $\bhat{\O}{}^1,\dots,\bhat{\O}{}^\Kk$. Specifically, the $\j{}^{\rm th}$ column of $\bhat{\O}{}^\k \in \{0,1\}{}^{\d \times \n}$ will contain a $1$ in the rows indicated by $\bhat{\o}{}^\k$.
\item[(ii)]
\textbf{Complete.}
For each $\k$, complete $\X_{\bhat{\O}{}^\k}$ using your favorite \LRMC\ algorithm. Then compute a new estimate $\bhat{\U}{}^\k$ given by the leading $\r$ left singular vectors of the completion of $\X_{\bhat{\O}{}^\k}$.
\end{itemize}
The entire procedure is summarized in Algorithm \ref{ammcAlg}. In Appendix \ref{algorithmApx} we further discuss initialization, generalizations to noise and outliers, and other simple extensions to improve performance.

\begin{algorithm}[tb]
   \caption{Alternating Mixture Matrix Completion (AMMC).}
   \label{ammcAlg}
\begin{algorithmic}
   \STATE $\Scale[.7]{\gray{1}}$. {\bfseries input:} Partially observed data matrix $\X_\O$.
   \STATE $\Scale[.7]{\gray{2}}$. {\bfseries initialize:} Guess $\bhat{\U}{}^1,\dots,\bhat{\U}{}^\Kk \in \R{}^{\d \times \r}$.
   \REPEAT
   \STATE {\bfseries CLUSTER:}
   	\FOR{$\j=1,\dots,\n$, and $\k=1,\dots,\Kk$}
		\STATE $\Scale[.7]{\gray{3}}$. $\x_\o = \j{}^{\rm th}$ column of $\X_\O$.
		\STATE $\Scale[.7]{\gray{4}}$. {\em Erase} entries from $\o$ to obtain $\ups{}^\k \subset \o$ indicating entries likely to correspond to $\X{}^\k$.
		\STATE $\Scale[.7]{\gray{5}}$. Estimate coefficient of $\j{}^{\rm th}$ column of $\X{}^\k$:
			\begin{align*}
			\bhat{\btheta}{}^\k \ = \ (\bhat{\U}{}^{\k\T}_{\ups{}^\k} \bhat{\U}{}^\k_{\ups{}^\k}){}^{-1}
				\bhat{\U}{}^{\k\T}_{\ups{}^\k} \x_{\ups{}^\k}.
			\end{align*}
		\STATE $\Scale[.7]{\gray{6}}$. The $\j{}^{\rm th}$ column of $\bhat{\X}{}^\k_\O$ is given by
			$\bhat{\x}{}^\k_\o \ = \ \bhat{\U}{}^\k_\o \bhat{\btheta}{}^\k$.
	\ENDFOR
	\STATE $\Scale[.7]{\gray{7}}$. Cluster the entries of $\X_\O$ according to their closest match among $\bhat{\X}{}^1_\O,\dots,\bhat{\X}{}^\Kk_\O$
	\STATE $\Scale[.7]{\white{7..}}$ to produce $\bhat{\O}{}^1,\dots,\bhat{\O}{}^\Kk$.
   \STATE {\bfseries COMPLETE:}
   	\FOR{$\k=1,\dots,\Kk$}
		\STATE $\Scale[.7]{\gray{8}}$. Complete $\X_{\bhat{\O}^\k}$ using \LRMC\ to obtain $\bhat{\X}{}^\k$.
		\STATE $\Scale[.7]{\gray{9}}$. $\bhat{\U}{}^\k=$ leading $\r$ singular vectors of $\bhat{\X}{}^\k$.
	\ENDFOR
   \UNTIL{convergence.}
   \STATE $\Scale[.7]{\gray{10}}$. {\bfseries output:} Completed matrices $\bhat{\X}{}^1,\dots,\bhat{\X}{}^\Kk$.
\end{algorithmic}
\end{algorithm}

\section{Experiments}
\label{experimentsSec}
\subsection{Simulations}
\begin{figure}
\centering
\Scale[.85]{\input{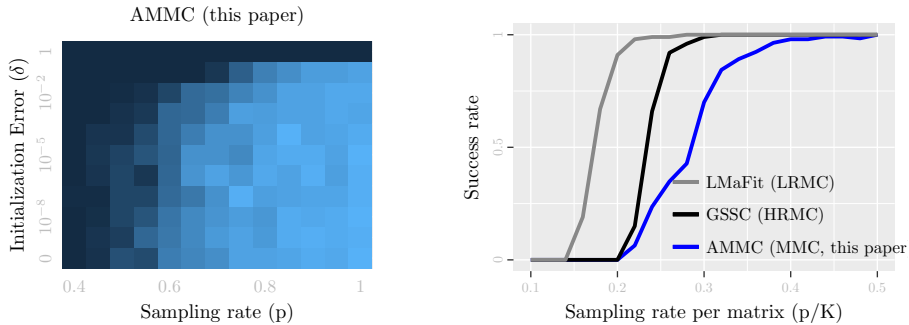}}
\hfill
\Scale[.85]{
\begin{tikzpicture}[x=1pt,y=1pt]
\definecolor{fillColor}{RGB}{255,255,255}
\path[use as bounding box,fill=fillColor,fill opacity=0.00] (0,0) rectangle (202.36,144.54);
\begin{scope}
\path[clip] (  0.00,  0.00) rectangle (202.36,144.54);
\definecolor{drawColor}{RGB}{255,255,255}
\definecolor{fillColor}{RGB}{255,255,255}

\path[draw=drawColor,line width= 0.6pt,line join=round,line cap=round,fill=fillColor] (  0.00,  0.00) rectangle (202.36,144.54);
\end{scope}
\begin{scope}
\path[clip] ( 27.23, 28.75) rectangle (196.86,139.04);
\definecolor{fillColor}{gray}{0.92}

\path[fill=fillColor] ( 27.23, 28.75) rectangle (196.86,139.04);
\definecolor{drawColor}{RGB}{255,255,255}

\path[draw=drawColor,line width= 0.3pt,line join=round] ( 27.23, 58.83) --
	(196.86, 58.83);

\path[draw=drawColor,line width= 0.3pt,line join=round] ( 27.23,108.96) --
	(196.86,108.96);

\path[draw=drawColor,line width= 0.3pt,line join=round] ( 54.21, 28.75) --
	( 54.21,139.04);

\path[draw=drawColor,line width= 0.3pt,line join=round] ( 92.76, 28.75) --
	( 92.76,139.04);

\path[draw=drawColor,line width= 0.3pt,line join=round] (131.32, 28.75) --
	(131.32,139.04);

\path[draw=drawColor,line width= 0.3pt,line join=round] (169.87, 28.75) --
	(169.87,139.04);

\path[draw=drawColor,line width= 0.6pt,line join=round] ( 27.23, 33.77) --
	(196.86, 33.77);

\path[draw=drawColor,line width= 0.6pt,line join=round] ( 27.23, 83.90) --
	(196.86, 83.90);

\path[draw=drawColor,line width= 0.6pt,line join=round] ( 27.23,134.03) --
	(196.86,134.03);

\path[draw=drawColor,line width= 0.6pt,line join=round] ( 34.94, 28.75) --
	( 34.94,139.04);

\path[draw=drawColor,line width= 0.6pt,line join=round] ( 73.49, 28.75) --
	( 73.49,139.04);

\path[draw=drawColor,line width= 0.6pt,line join=round] (112.04, 28.75) --
	(112.04,139.04);

\path[draw=drawColor,line width= 0.6pt,line join=round] (150.59, 28.75) --
	(150.59,139.04);

\path[draw=drawColor,line width= 0.6pt,line join=round] (189.15, 28.75) --
	(189.15,139.04);
\definecolor{drawColor}{RGB}{0,0,255}

\path[draw=drawColor,line width= 1.7pt,line join=round] ( 34.94, 33.77) --
	( 42.65, 33.77) --
	( 50.36, 33.77) --
	( 58.07, 33.77) --
	( 65.78, 33.77) --
	( 73.49, 33.77) --
	( 81.20, 40.18) --
	( 88.91, 57.43) --
	( 96.62, 68.66) --
	(104.33, 76.68) --
	(112.04,103.95) --
	(119.75,118.39) --
	(127.46,123.20) --
	(135.17,126.41) --
	(142.88,130.42) --
	(150.59,132.02) --
	(158.30,132.02) --
	(166.01,133.22) --
	(173.72,133.22) --
	(181.44,132.42) --
	(189.15,134.03);
\definecolor{drawColor}{RGB}{0,0,0}

\path[draw=drawColor,line width= 1.7pt,line join=round] ( 34.94, 33.77) --
	( 42.65, 33.77) --
	( 50.36, 33.77) --
	( 58.07, 33.77) --
	( 65.78, 33.77) --
	( 73.49, 33.77) --
	( 81.20, 48.80) --
	( 88.91, 99.94) --
	( 96.62,126.01) --
	(104.33,130.02) --
	(112.04,133.02) --
	(119.75,134.03) --
	(127.46,134.03) --
	(135.17,134.03) --
	(142.88,134.03) --
	(150.59,134.03) --
	(158.30,134.03) --
	(166.01,134.03) --
	(173.72,134.03) --
	(181.44,134.03) --
	(189.15,134.03);
\definecolor{drawColor}{RGB}{136,136,136}

\path[draw=drawColor,line width= 1.7pt,line join=round] ( 34.94, 33.77) --
	( 42.65, 33.77) --
	( 50.36, 33.77) --
	( 58.07, 52.82) --
	( 65.78,100.94) --
	( 73.49,125.00) --
	( 81.20,132.02) --
	( 88.91,133.02) --
	( 96.62,133.02) --
	(104.33,134.03) --
	(112.04,134.03) --
	(119.75,134.03) --
	(127.46,134.03) --
	(135.17,134.03) --
	(142.88,134.03) --
	(150.59,134.03) --
	(158.30,134.03) --
	(166.01,134.03) --
	(173.72,134.03) --
	(181.44,134.03) --
	(189.15,134.03);
\end{scope}
\begin{scope}
\path[clip] (  0.00,  0.00) rectangle (202.36,144.54);
\definecolor{drawColor}{RGB}{190,190,190}

\node[text=drawColor,rotate= 90.00,anchor=base,inner sep=0pt, outer sep=0pt, scale=  0.62] at ( 22.28, 33.77) {0};

\node[text=drawColor,rotate= 90.00,anchor=base,inner sep=0pt, outer sep=0pt, scale=  0.62] at ( 22.28, 83.90) {0.5};

\node[text=drawColor,rotate= 90.00,anchor=base,inner sep=0pt, outer sep=0pt, scale=  0.62] at ( 22.28,134.03) {1};
\end{scope}
\begin{scope}
\path[clip] (  0.00,  0.00) rectangle (202.36,144.54);
\definecolor{drawColor}{gray}{0.20}

\path[draw=drawColor,line width= 0.6pt,line join=round] ( 24.48, 33.77) --
	( 27.23, 33.77);

\path[draw=drawColor,line width= 0.6pt,line join=round] ( 24.48, 83.90) --
	( 27.23, 83.90);

\path[draw=drawColor,line width= 0.6pt,line join=round] ( 24.48,134.03) --
	( 27.23,134.03);
\end{scope}
\begin{scope}
\path[clip] (  0.00,  0.00) rectangle (202.36,144.54);
\definecolor{drawColor}{gray}{0.20}

\path[draw=drawColor,line width= 0.6pt,line join=round] ( 34.94, 26.00) --
	( 34.94, 28.75);

\path[draw=drawColor,line width= 0.6pt,line join=round] ( 73.49, 26.00) --
	( 73.49, 28.75);

\path[draw=drawColor,line width= 0.6pt,line join=round] (112.04, 26.00) --
	(112.04, 28.75);

\path[draw=drawColor,line width= 0.6pt,line join=round] (150.59, 26.00) --
	(150.59, 28.75);

\path[draw=drawColor,line width= 0.6pt,line join=round] (189.15, 26.00) --
	(189.15, 28.75);
\end{scope}
\begin{scope}
\path[clip] (  0.00,  0.00) rectangle (202.36,144.54);
\definecolor{drawColor}{RGB}{190,190,190}

\node[text=drawColor,anchor=base,inner sep=0pt, outer sep=0pt, scale=  0.62] at ( 34.94, 19.56) {0.1};

\node[text=drawColor,anchor=base,inner sep=0pt, outer sep=0pt, scale=  0.62] at ( 73.49, 19.56) {0.2};

\node[text=drawColor,anchor=base,inner sep=0pt, outer sep=0pt, scale=  0.62] at (112.04, 19.56) {0.3};

\node[text=drawColor,anchor=base,inner sep=0pt, outer sep=0pt, scale=  0.62] at (150.59, 19.56) {0.4};

\node[text=drawColor,anchor=base,inner sep=0pt, outer sep=0pt, scale=  0.62] at (189.15, 19.56) {0.5};
\end{scope}
\begin{scope}
\path[clip] (  0.00,  0.00) rectangle (202.36,144.54);
\definecolor{drawColor}{RGB}{0,0,0}

\node[text=drawColor,anchor=base,inner sep=0pt, outer sep=0pt, scale=  0.88] at (112.04,  8.00) {Sampling rate per matrix (${\rm p}/{\rm K}$)};
\end{scope}
\begin{scope}
\path[clip] (  0.00,  0.00) rectangle (202.36,144.54);
\definecolor{drawColor}{RGB}{0,0,0}

\node[text=drawColor,rotate= 90.00,anchor=base,inner sep=0pt, outer sep=0pt, scale=  0.88] at ( 11.56, 83.90) {Success rate};
\end{scope}
\begin{scope}
\path[clip] (  0.00,  0.00) rectangle (202.36,144.54);

\path[] ( 91.01, 26.67) rectangle (211.10, 92.60);
\end{scope}
\begin{scope}
\path[clip] (  0.00,  0.00) rectangle (202.36,144.54);

\path[] ( 96.70, 61.27) rectangle (111.16, 75.72);
\end{scope}
\begin{scope}
\path[clip] (  0.00,  0.00) rectangle (202.36,144.54);
\definecolor{drawColor}{RGB}{136,136,136}

\path[draw=drawColor,line width= 1.7pt,line join=round] ( 98.15, 68.49) -- (109.71, 68.49);
\end{scope}
\begin{scope}
\path[clip] (  0.00,  0.00) rectangle (202.36,144.54);
\definecolor{drawColor}{RGB}{136,136,136}

\path[draw=drawColor,line width= 1.7pt,line join=round] ( 98.15, 68.49) -- (109.71, 68.49);
\end{scope}
\begin{scope}
\path[clip] (  0.00,  0.00) rectangle (202.36,144.54);
\definecolor{drawColor}{RGB}{136,136,136}

\path[draw=drawColor,line width= 1.7pt,line join=round] ( 98.15, 68.49) -- (109.71, 68.49);
\end{scope}
\begin{scope}
\path[clip] (  0.00,  0.00) rectangle (202.36,144.54);

\path[] ( 96.70, 46.81) rectangle (111.16, 61.27);
\end{scope}
\begin{scope}
\path[clip] (  0.00,  0.00) rectangle (202.36,144.54);
\definecolor{drawColor}{RGB}{0,0,0}

\path[draw=drawColor,line width= 1.7pt,line join=round] ( 98.15, 54.04) -- (109.71, 54.04);
\end{scope}
\begin{scope}
\path[clip] (  0.00,  0.00) rectangle (202.36,144.54);
\definecolor{drawColor}{RGB}{0,0,0}

\path[draw=drawColor,line width= 1.7pt,line join=round] ( 98.15, 54.04) -- (109.71, 54.04);
\end{scope}
\begin{scope}
\path[clip] (  0.00,  0.00) rectangle (202.36,144.54);
\definecolor{drawColor}{RGB}{0,0,0}

\path[draw=drawColor,line width= 1.7pt,line join=round] ( 98.15, 54.04) -- (109.71, 54.04);
\end{scope}
\begin{scope}
\path[clip] (  0.00,  0.00) rectangle (202.36,144.54);

\path[] ( 96.70, 32.36) rectangle (111.16, 46.81);
\end{scope}
\begin{scope}
\path[clip] (  0.00,  0.00) rectangle (202.36,144.54);
\definecolor{drawColor}{RGB}{0,0,255}

\path[draw=drawColor,line width= 1.7pt,line join=round] ( 98.15, 39.58) -- (109.71, 39.58);
\end{scope}
\begin{scope}
\path[clip] (  0.00,  0.00) rectangle (202.36,144.54);
\definecolor{drawColor}{RGB}{0,0,255}

\path[draw=drawColor,line width= 1.7pt,line join=round] ( 98.15, 39.58) -- (109.71, 39.58);
\end{scope}
\begin{scope}
\path[clip] (  0.00,  0.00) rectangle (202.36,144.54);
\definecolor{drawColor}{RGB}{0,0,255}

\path[draw=drawColor,line width= 1.7pt,line join=round] ( 98.15, 39.58) -- (109.71, 39.58);
\end{scope}
\begin{scope}
\path[clip] (  0.00,  0.00) rectangle (202.36,144.54);
\definecolor{drawColor}{RGB}{0,0,0}

\node[text=drawColor,anchor=base west,inner sep=0pt, outer sep=0pt, scale=  0.77] at (112.97, 65.84) {LMaFit (LRMC)};
\end{scope}
\begin{scope}
\path[clip] (  0.00,  0.00) rectangle (202.36,144.54);
\definecolor{drawColor}{RGB}{0,0,0}

\node[text=drawColor,anchor=base west,inner sep=0pt, outer sep=0pt, scale=  0.77] at (112.97, 51.39) {GSSC (HRMC)};
\end{scope}
\begin{scope}
\path[clip] (  0.00,  0.00) rectangle (202.36,144.54);
\definecolor{drawColor}{RGB}{0,0,0}

\node[text=drawColor,anchor=base west,inner sep=0pt, outer sep=0pt, scale=  0.77] at (112.97, 36.93) {AMMC (MMC, this paper)};
\end{scope}
\end{tikzpicture}}
\caption{{\bf Left:} Success rate (average over $100$ trials) of \AMMC\ as a function of the fraction of observed entries $\p$ and the distance $\deltaa$ between the {\em true} subspaces $\U{}^\k$ and their initial estimates. Lightest represents $100\%$ success rate; darkest represents $0\%$. {\bf Right:} Comparison of state-of-the-art algorithms for \LRMC, \HRMC, and \MMC\ (in their respective settings; see Figure \ref{generalizationMMCFig}).  The performance of \AMMC\ (in the more difficult problem of \MMC) is comparable to the performance of state-of-the-art algorithms in the simpler problems of \LRMC\ and \HRMC.}
\label{ammcFig}
\end{figure}

We first present a series of synthetic experiments to study the performance of \AMMC\ (Algorithm \ref{ammcAlg}). In our simulations we first generate matrices $\U{}^\k \in \R{}^{\d \times \r}$ and \phantomsection\label{bThetaDef}$\bTheta{}^\k \in \R{}^{\r \times \n}$ with i.i.d.~$\mathscr{N}(0,1)$ entries to use as bases and coefficients of the low-rank matrices in the mixture, i.e., $\X{}^\k=\U{}^\k \bTheta{}^\k \in \R{}^{\d \times \n}$. Here $\d=\n=100$, $\r=5$ and $\Kk=2$. With probability $(1-\p)$, the $(\i,\j)^{\rm th}$ entry of $\X_\O$ will be missing, and with probability $\nicefrac{\p}{\Kk}$ it will be equal to the corresponding entry in $\X{}^\k$. Recall that similar to EM and other alternating approaches, \AMMC\ depends on initialization. Hence, we study the performance of \AMMC\ as a function of both $\p$ and the distance \phantomsection\label{deltaaDef}$\deltaa \in [0,1]$ between $\{\U{}^\k\}$ and their initial estimates (measured as the normalized Frobenius norm of the difference between their projection operators). We measure accuracy using the normalized Frobenius norm of the difference between each $\X{}^\k$ and its completion. We considered a success if this quantity was below $10{}^{-8}$. The results of $100$ trials are summarized in Figure \ref{ammcFig}.

Notice that the performance of \AMMC\ decays nicely with the distance $\deltaa$ between the {\em true} subspaces $\U{}^\k$ and their initial estimates. We can see this type of behavior in similar state-of-the-art alternating algorithms for the simpler problem of \HRMC\ \cite{gssc}. Since \MMC\ is highly non-convex, it is not surprising that if the initial estimates are poor (far from the truth), then \AMMC\ may converge to a local minimum.  Similarly, the performance of \AMMC\ decays nicely with the fraction of observed entries $\p$. Notice that even if $\X$ is fully observed ($\p=1$), if the initial estimates are very far from the true subspaces ($\deltaa=1$), then \AMMC\ performs poorly. This shows, consistent with our discussing in Remark \ref{difficultyRmk}, that in practice \MMC\ is a challenging problem even if $\X$ is fully observed.
Hence, it is quite remarkable that \AMMC\ works most of the time with as little as $\p \approx 0.6$, corresponding to observing $\approx 0.3$ of the entries in each $\X{}^\k$. To put this under perspective, notice (Figure \ref{ammcFig}) that this is comparable the amount of missing data tolerated by GSSC \cite{gssc} and LMaFit \cite{lmafit}, which are state-of-the-art for the simpler problems of \HRMC\ (special case of \MMC\ where all entries in each column of $\X$ correspond to the same $\X{}^\k$) and \LRMC\ (special case where there is only one $\X{}^\k$). To obtain Figure \ref{ammcFig} we replicated the same setup as above, but with data generated according to the \HRMC\ and \LRMC\ models. Hence, we conclude that the performance of \AMMC\ (in the more difficult problem of \MMC) is comparable to the performance of state-of-the-art algorithms for the much simpler problems of \HRMC\ and \LRMC.

We point out that according to Theorems \ref{mainThm} and \ref{probabilityThm}, \MMC\ is theoretically possible with $\p \geq \nicefrac{1}{2}$. However, we can see that (even if $\U{}^1,\dots,\U{}^\Kk$ are known, corresponding to $\deltaa=0$ in Figure \ref{ammcFig}) the performance of \AMMC\ is quite poor if $\p<0.6$. This shows two things: (i) \MMC\ is challenging even if $\U{}^1,\dots,\U{}^\Kk$ are known (as discussed in Remark \ref{difficultyRmk}), and (ii) there is a gap between what is information-theoretically possible and what is currently possible in practice (with \AMMC). In future work we will explore algorithms that can approach the information-theoretic limits.


\subsection{Real Data: Face Clustering and Inpainting}
\begin{figure}
\centering
\begin{tabular}{cc}
\rotatebox{90}{\hspace{.1cm} Mixture \hspace{.2cm}} &
\includegraphics[height=1.75cm]{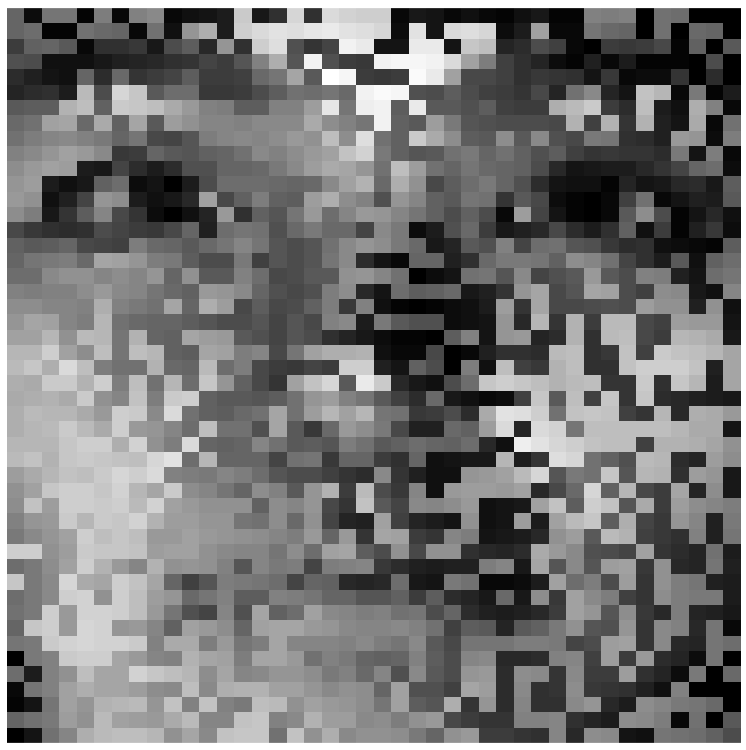}
\includegraphics[height=1.75cm]{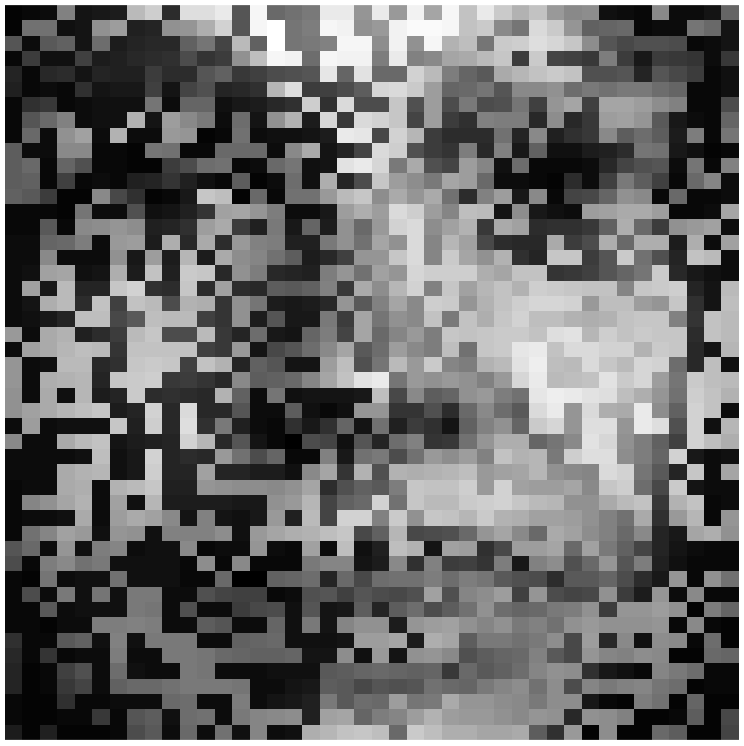}
\includegraphics[height=1.75cm]{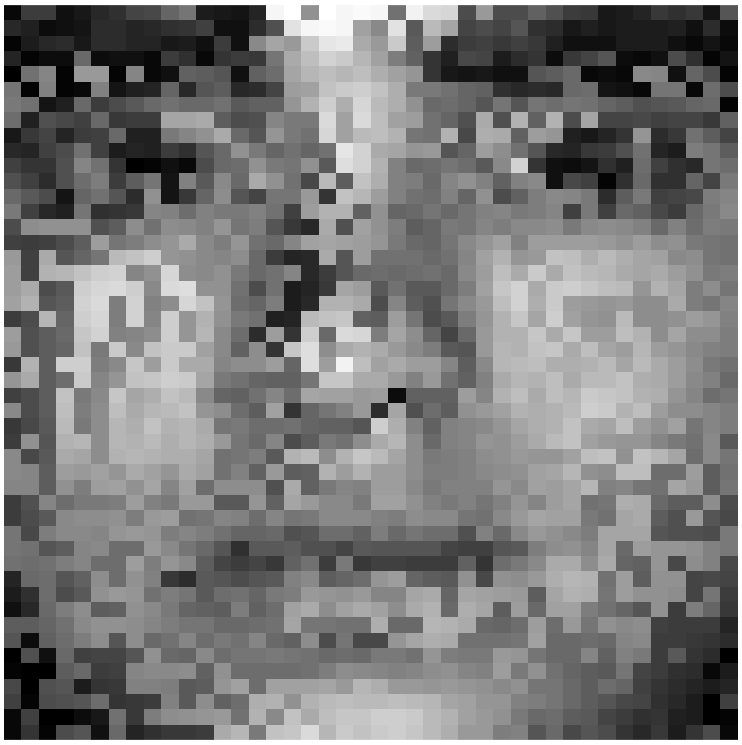}
\includegraphics[height=1.75cm]{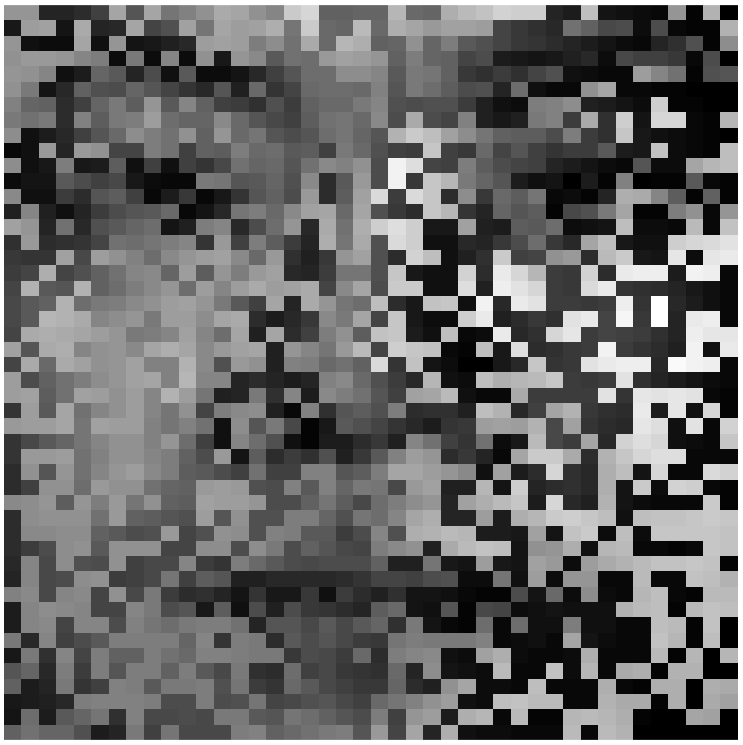}
\includegraphics[height=1.75cm]{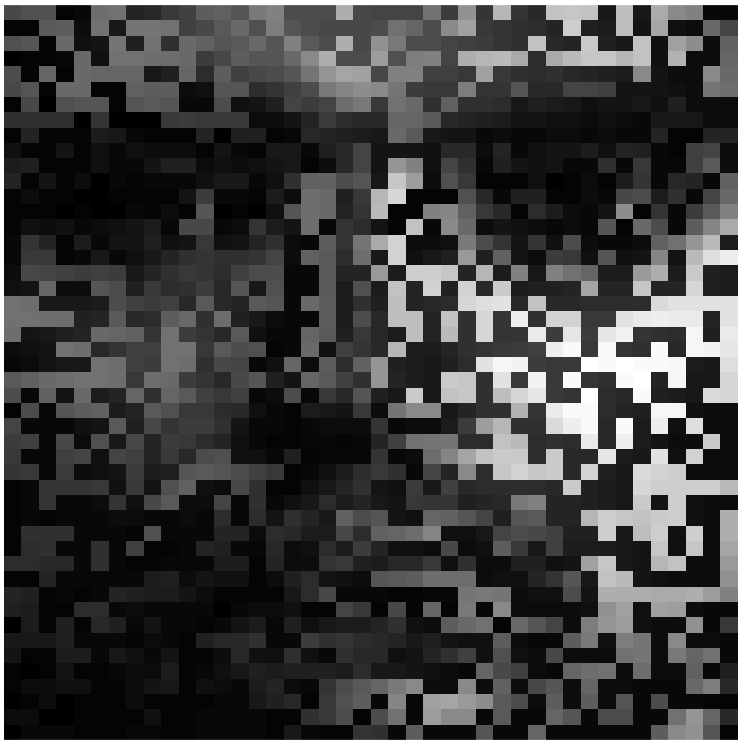}
\includegraphics[height=1.75cm]{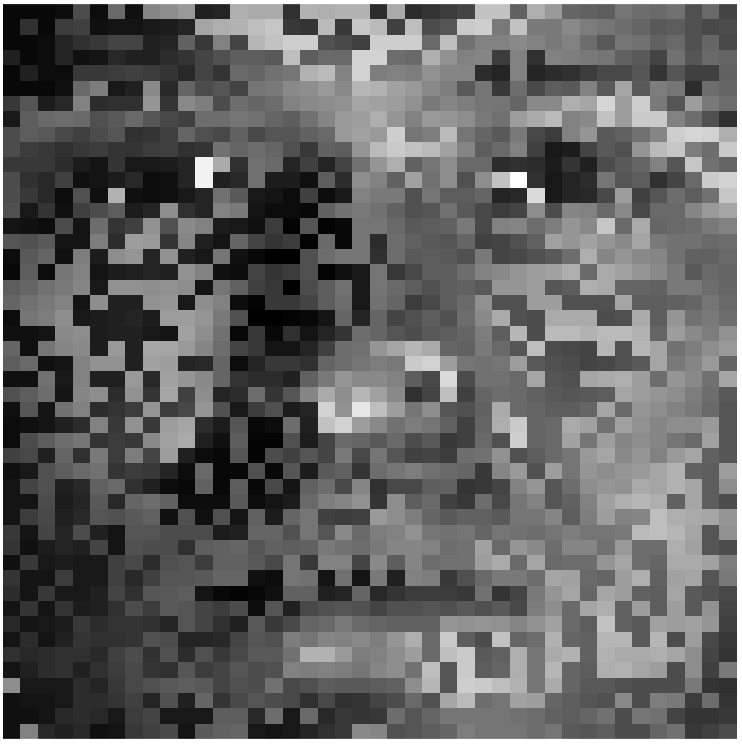} \\
\multirow{-2}[13]{*}{\rotatebox{90}{\------ Reconstructions \------\phantom{\----}}} &
\includegraphics[height=1.75cm]{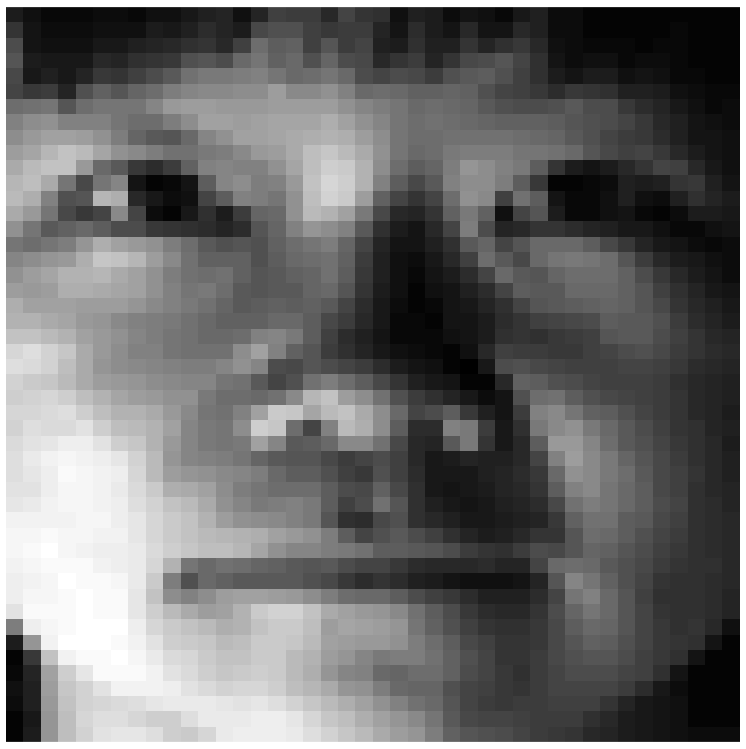}
\includegraphics[height=1.75cm]{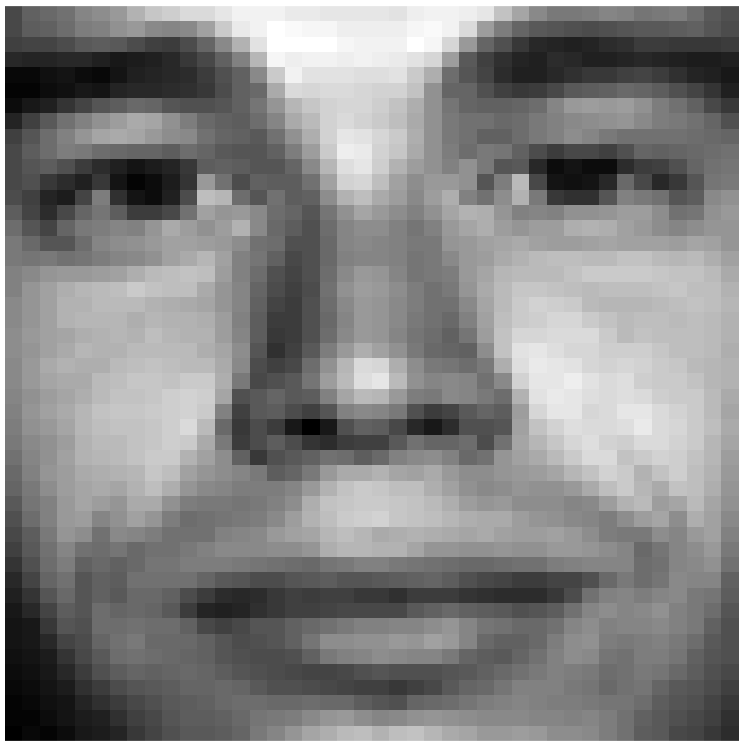}
\includegraphics[height=1.75cm]{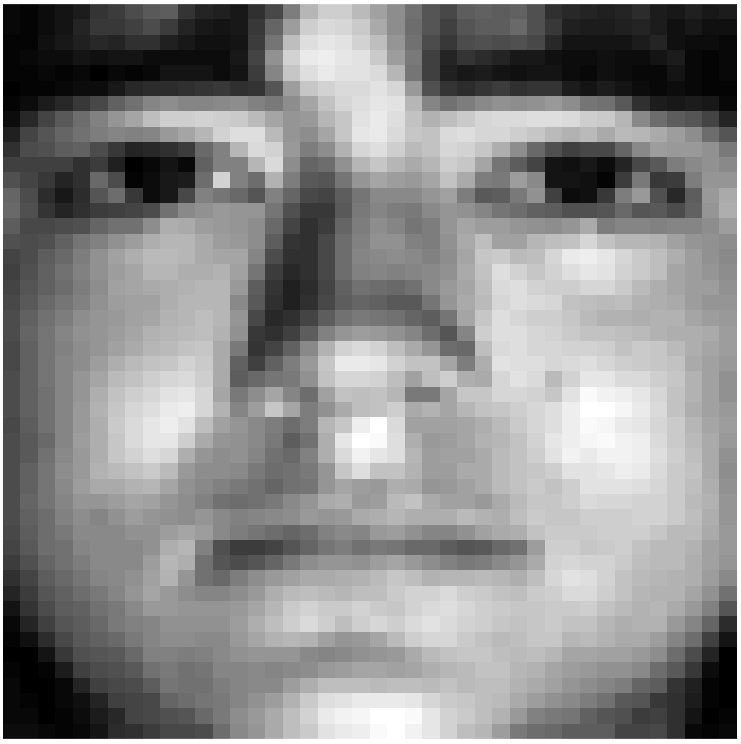}
\includegraphics[height=1.75cm]{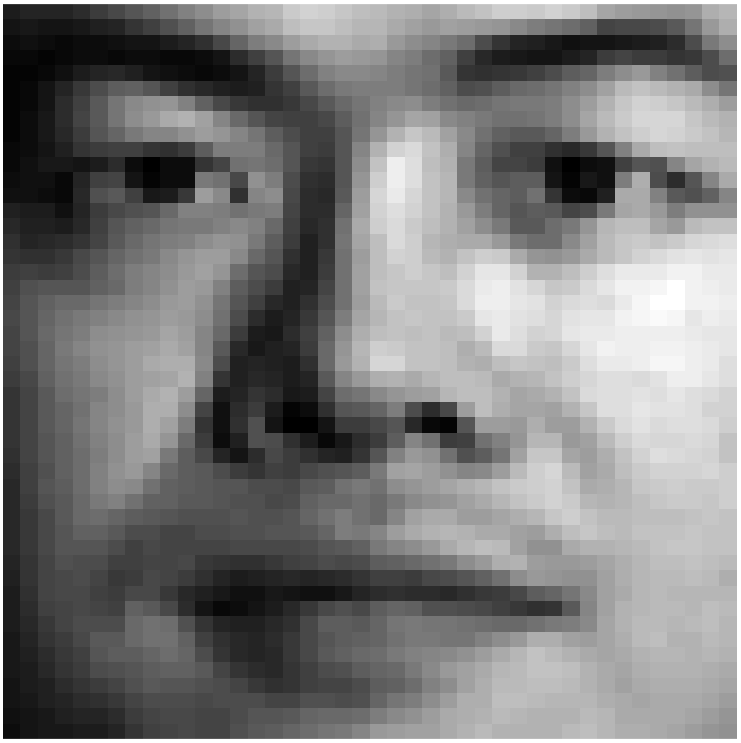}
\includegraphics[height=1.75cm]{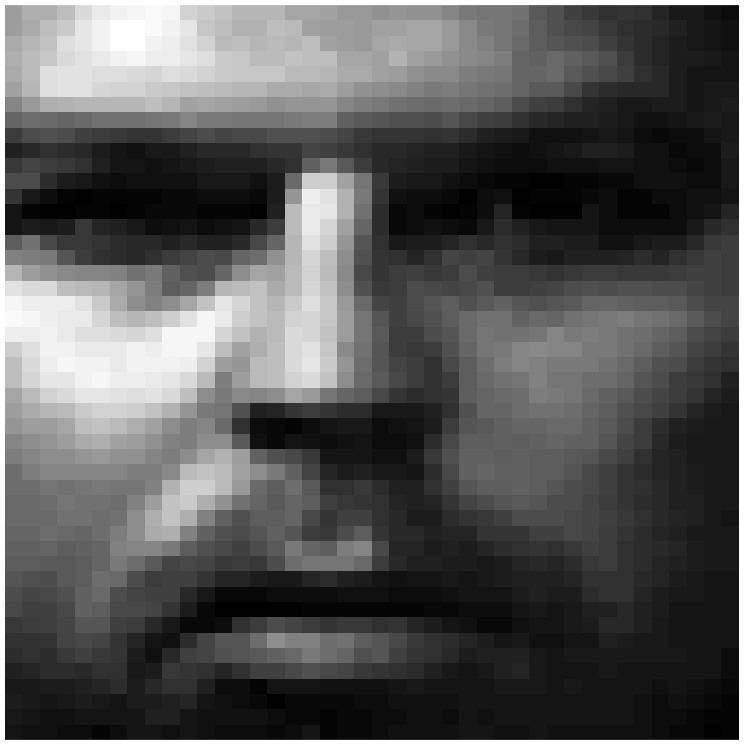}
\includegraphics[height=1.75cm]{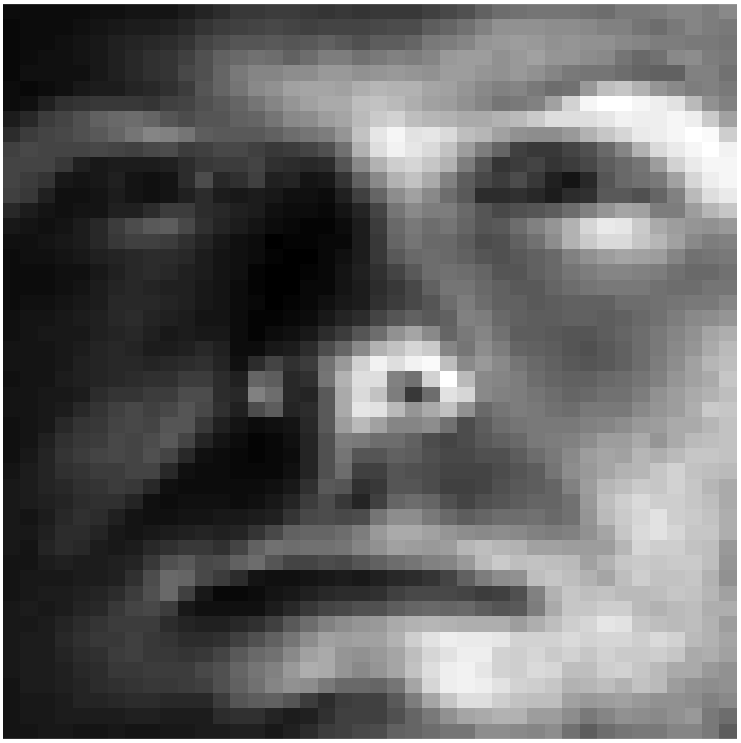} \\
&
\includegraphics[height=1.75cm]{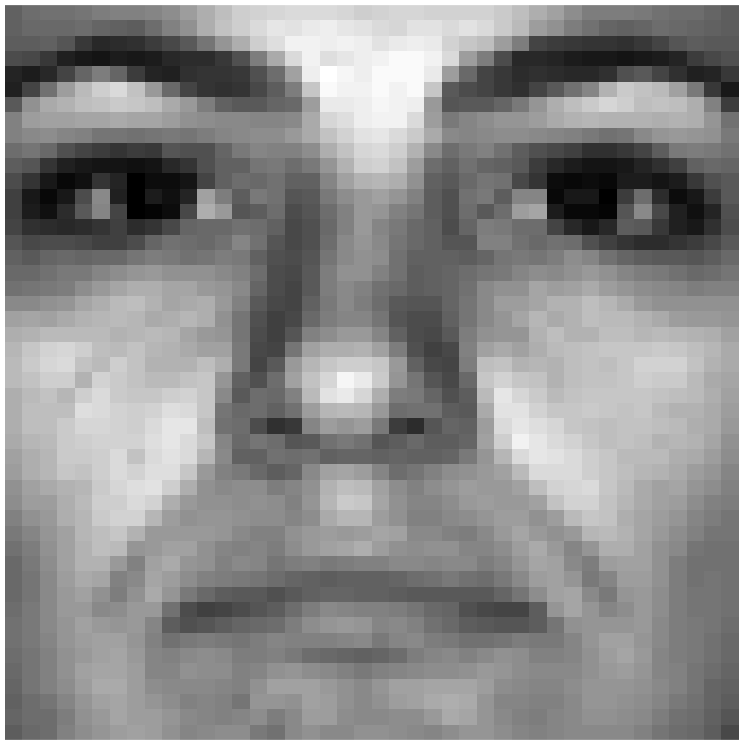}
\includegraphics[height=1.75cm]{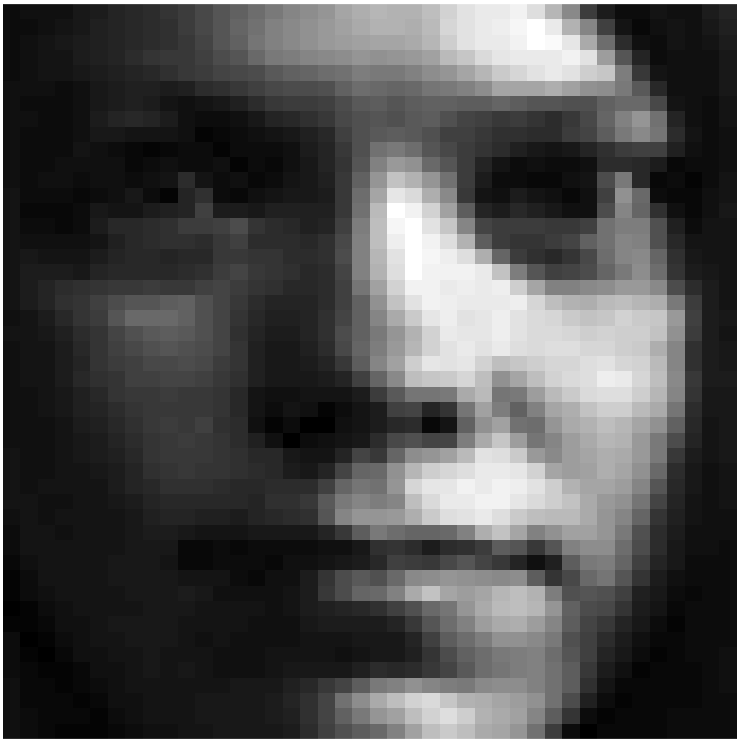}
\includegraphics[height=1.75cm]{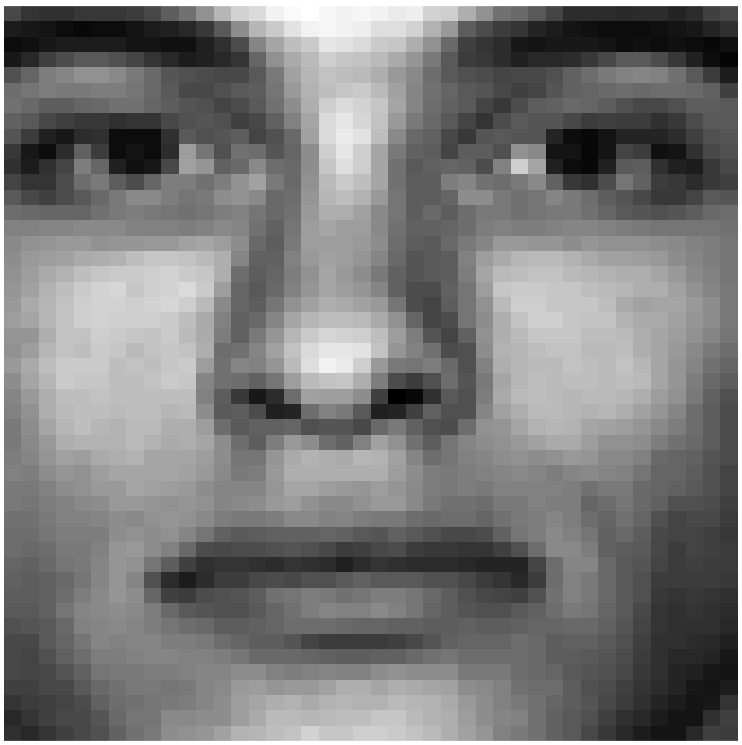}
\includegraphics[height=1.75cm]{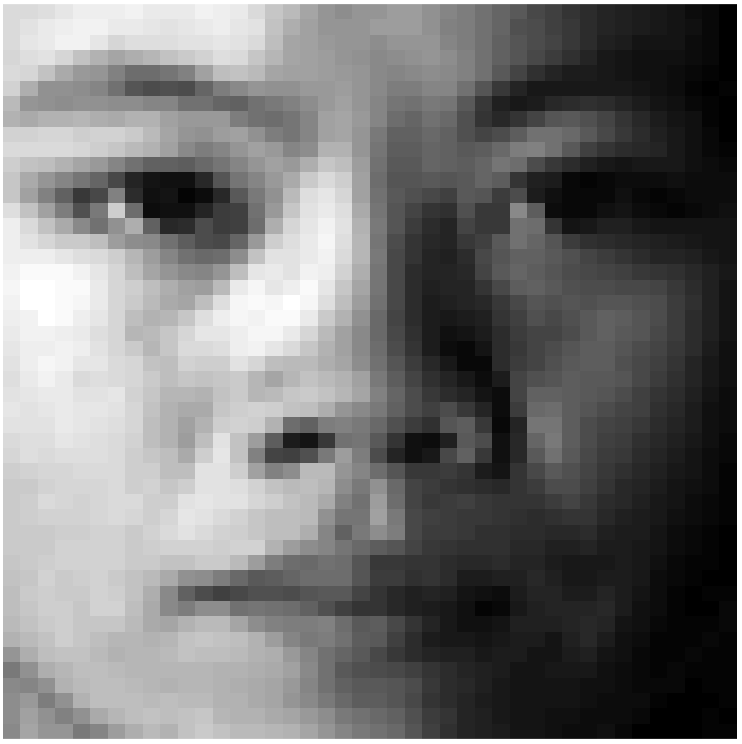}
\includegraphics[height=1.75cm]{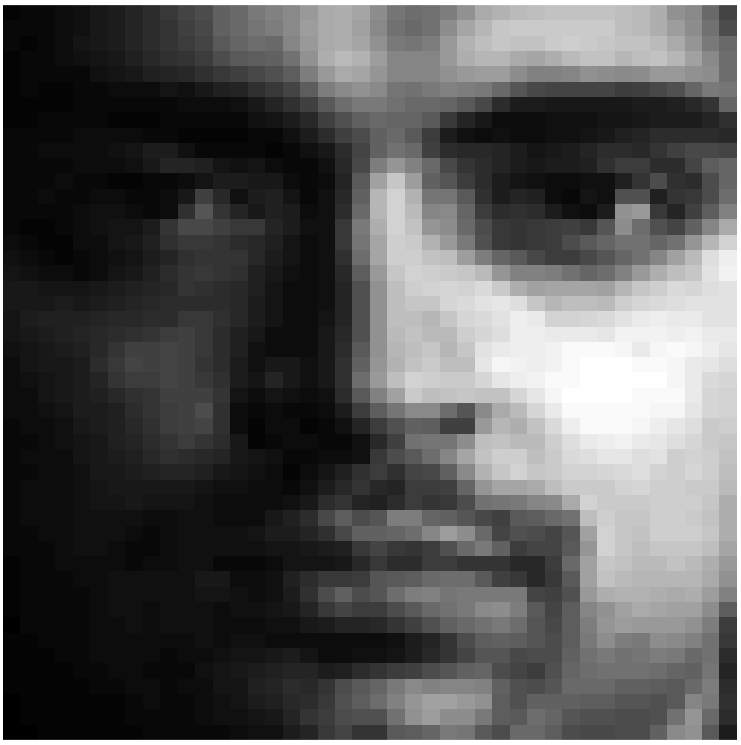}
\includegraphics[height=1.75cm]{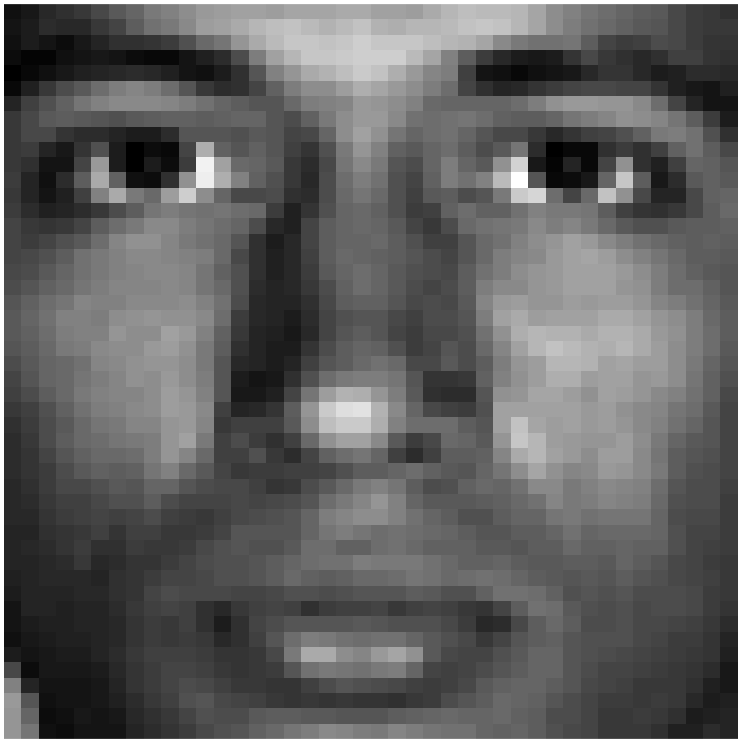}
\end{tabular}
\caption{{\bf Top:} Mixture matrix $\X$, containing pixels from two face images. {\bf Bottom 2:} Low-rank matrices $\bhat{\X}{}^1$ and $\bhat{\X}{}^2$ recovered from $\X$.}
\label{yaleFig}
\end{figure}

It is well-known that images of an individual's face are approximately low-rank \cite{lambertian}. Natural images, however, usually contain faces of multiple individuals, often partially occluding each other, resulting in a mixture of low-rank matrices. In this experiment we demonstrate the power of \MMC\ in two tasks: first, classifying partially occluded faces in an image, and second, image inpainting \cite{inpainting}. To this end, we use the Yale B dataset \cite{yale}, containing $2432$ photos of $38$ subjects (64 photos per subject), each photo of size $48 \times 42$. We randomly select two subjects, and vectorize and concatenate their images to obtain two approximately rank-$10$ matrices $\X{}^1,\X{}^2 \in \R^{2016 \times 64}$. Next we combine them into $\X \in \R^{2016 \times 64}$, whose each entry is equal to the corresponding entry in $\X{}^1$ or $\X{}^2$ with equal probability. This way, each column of $\X$ contains a mixed image with pixels from multiple individuals. We aim at two goals: (i) classify the entries in $\X$ according to $\X{}^1$ and $\X{}^2$, which in turn means locating and classifying the face of each individual in each image, and (ii) recover $\X{}^1$ and $\X{}^2$ from $\X$, thus reconstructing the unobserved pixels in each image (inpainting). We repeat this experiment $30$ times using \AMMC\ (with gaussian random initialization, known to produce near-orthogonal subspaces with high probability), obtaining a pixel classification error of $2.98\%$, and a reconstruction error of $4.1\%$, which is remarkable in light that the ideal rank-$10$ approximation (no mixture, and full data) achieves $1.8\%$. Figure \ref{yaleFig} shows a few examples. Notice that in this case we cannot compare against other methods, as \AMMC\ is the first, and currently the only method for \MMC.

\subsection{Real Data: \MMC\ for Background Segmentation}
\begin{figure}
\centering
\begin{tabular}{ccc}
\hspace{.4cm} Original &
\hspace{.4cm} \--------- Robust \PCA\ \--------- &
\hspace{.2cm} \------ \AMMC\ (this paper) \------
\end{tabular} \\
\includegraphics[width=2.3cm]{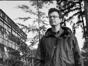}
\includegraphics[width=2.3cm]{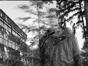} 
\includegraphics[width=2.3cm]{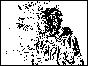}
\includegraphics[width=2.3cm]{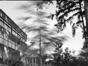}
\includegraphics[width=2.3cm]{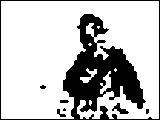} \\
\includegraphics[width=2.3cm]{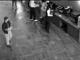}
\includegraphics[width=2.3cm]{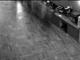} 
\includegraphics[width=2.3cm]{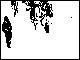}
\includegraphics[width=2.3cm]{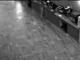}
\includegraphics[width=2.3cm]{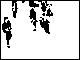} \\
\includegraphics[width=2.3cm]{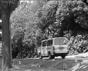}
\includegraphics[width=2.3cm]{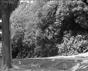}
\includegraphics[width=2.3cm]{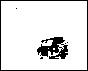}
\includegraphics[width=2.3cm]{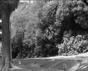}
\includegraphics[width=2.3cm]{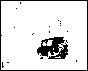} \\
\includegraphics[width=2.3cm]{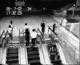}
\includegraphics[width=2.3cm]{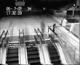}
\includegraphics[width=2.3cm]{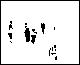}
\includegraphics[width=2.3cm]{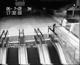}
\includegraphics[width=2.3cm]{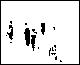} \\
\includegraphics[width=2.3cm]{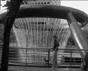}
\includegraphics[width=2.3cm]{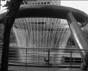}
\includegraphics[width=2.3cm]{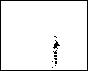}
\includegraphics[width=2.3cm]{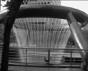}
\includegraphics[width=2.3cm]{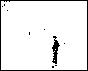} \\
\includegraphics[width=2.3cm]{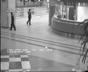}
\includegraphics[width=2.3cm]{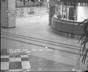}
\includegraphics[width=2.3cm]{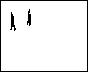}
\includegraphics[width=2.3cm]{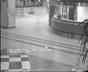}
\includegraphics[width=2.3cm]{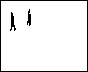} \\
\includegraphics[width=2.3cm]{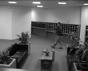}
\includegraphics[width=2.3cm]{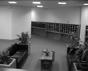}
\includegraphics[width=2.3cm]{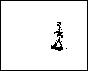}
\includegraphics[width=2.3cm]{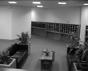}
\includegraphics[width=2.3cm]{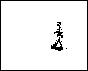} \\
\includegraphics[width=2.3cm]{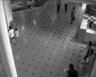}
\includegraphics[width=2.3cm]{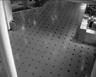}
\includegraphics[width=2.3cm]{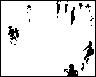}
\includegraphics[width=2.3cm]{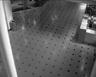}
\includegraphics[width=2.3cm]{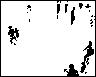}
\caption{Video frames segmented into background and foreground using robust \PCA\ (displaying the best results amongst \cite{alm, almNIPS, brpca, rosl, iht}) and \AMMC.}
\label{backgroundFig}
\end{figure}

As discussed in Section \ref{applicationsSec}, robust \PCA\ models a video as the superposition of a low-rank background plus a sparse foreground with no structure. \MMC\ brings more flexibility, allowing multiple low-rank matrices to model background, structured foreground objects (sparse or abundant) and illumination artifacts, while at the same time also accounting for outliers (the entries/pixels that were assigned to no matrix in the mixture). In fact, contrary to robust \PCA, \MMC\ allows a very large (even dominant) fraction of outliers. In this experiment we test \AMMC\ in the task of background segmentation, using the Wallflower \cite{wallflower} and the I2R \cite{background} datasets, containing videos of traffic cameras, lobbies, and pedestrians in the street. For each video, we compare \AMMC\ (with gaussian random initialization) against the best result amongst the following state-of-the-art algorithms for robust \PCA: \cite{alm, almNIPS, brpca, rosl, iht}. We chose these methods based on the comprehensive review in \cite{review}, and previous reports \cite{robustpca, survey, rpcaWebsite} indicating that these algorithms typically performed as well or better than several others, including \cite{alternating, apg}. Figure \ref{backgroundFig} contains some results. In most cases, both robust \PCA\ and \AMMC\ perform quite similarly. However, in one case (first row) \AMMC\ achieves $87.67\%$ segmentation accuracy (compared with the ground truth, manually segmented), while robust \PCA\ only achieves $74.88\%$ (Figure \ref{yaleFig}). Our hypothesis is that this is due to the large portion of outliers (foreground).  It is out of the scope of this paper, but of interest for future work, to collect real datasets with similar properties, where \AMMC\ can be further tested. We point out, however, that \AMMC\ is orders of magnitude slower than Robust \PCA. Our future work will also focus on developing faster methods for \MMC.

\newpage

\newpage
\appendix
\section{\MMC\ with Known Subspaces and Full Data}
\label{furtherApx}
Remark \ref{difficultyRmk} points out that if we knew the subspace(s) containing the columns in $\X$, then \LRMC\ and \HRMC\ become almost trivial problems, while \MMC\ still remains challenging. To see this, recall that by definition, the columns in a rank-$\r$ matrix lie in an $\r$-dimensional subspace. Recall that $\x_\o$ denotes the $\j{}^{\rm th}$ column of $\X_\O$, observed only on the entries indexed by $\o \subset \{1,\dots,\d\}$, and that $\U{}^\k \in \R{}^{\d \times \r}$ spans the subspace containing the columns of $\X{}^\k$.

Next suppose that all the entries in $\x_\o$ correspond to the same subspace (as would be the case in \HRMC\ and \LRMC), and that we know $\U{}^1,\dots,\U{}^\Kk$. Then we can project $\x_\o$ onto the subspaces generated by $\U{}_\o^1,\dots,\U{}_\o^\Kk$ to determine which subspace $\x_\o$ corresponds to. Say it is $\U{}^\k$. Then we can compute the coefficient of $\x_\o$ as $\btheta \ = \ (\U{}^{\k\T}_\o \U{}^\k_\o){}^{-1} \U{}^{\k\T}_\o \x_\o$. Since the coefficient of $\x$ is the same as the coefficient of $\x_\o$, we can recover $\x=\U{}^\k \btheta$.

In contrast, in \MMC\ the entries in $\x_\o$ may belong to multiple subspaces, and hence, even if we know $\U{}^1,\dots,\U{}^\Kk$, we cannot just project to identify the subspace corresponding to $\x_\o$ (if $\x_\o$ has entries from more than one subspace, it will not lie in {\em any} of the $\Kk$ subspaces). Hence, \MMC\ can be very challenging even if we know $\U{}^1,\dots,\U{}^\Kk$. This can be seen in our experiments. In particular pay attention to the bottom row in Figure \ref{ammcFig}, which shows the \MMC\ error when $\U{}^1,\dots,\U{}^\Kk$ are known.


Similarly, \MMC\ is difficult even if $\X$ is fully observed! To build some intuition, consider \HRMC. If no data is missing, \HRMC\ simplifies to \phantomsection\label{SCDef}{\em subspace clustering} (\SC) \cite{sc}, which has been studied extensively in recent years to produce theory and algorithms to handle gross errors \cite{liu1, liu2, mahdi, qu, peng}, noise \cite{wang}, privacy \cite{aarti} and data constraints \cite{hu}. Furthermore, the renowned state-of-the-art algorithm {\em sparse subspace clustering} \cite{ssc}, can efficiently, accurately and provably perform \SC. Hence, if $\X$ is fully observed, \HRMC\ is well understood.

In contrast, even if $\X$ is fully observed, \MMC\ remains \MMC, because we still do not know which entries belong together, and because for each entry in $\X$ that we observe, there are $\Kk-1$ that we do not. For example, if we observe an entry of $\X$ corresponding to $\X{}^1$, we still do not know that it belongs to $\X{}^1$, and we still need to recover the corresponding entries of $\X{}^2,\dots,\X{}^\Kk$. Furthermore, as we discussed above, and in Section \ref{algorithmSec}, even if $\U{}^1,\dots,\U{}^\Kk$ were known, identifying the entries that agree with the subspace is not a trivial problem. Hence, \MMC\ remains a challenging problem even with {\em full} data. This can be seen in our experiments. In particular pay attention to the last column in Figure \ref{ammcFig}, which shows the \MMC\ error when $\X$ is fully observed.

\section{Proofs}
\label{proofApx}
As discussed in Section \ref{problemSec}, the main subtlety in \MMC\ is that since we do not know a priori which entries of $\X_\O$ correspond to each $\X{}^\k$, there could arise {\em false} mixtures that agree with $\X_\O$. Fortunately, Theorem 3 in \cite{infoTheoretic} gives conditions to guarantee that a subset of entries correspond to the same $\X{}^\k$. We restate this result as the following Lemma, with some adaptations to our context.

\begin{myLemma}[Theorem 3 in \cite{infoTheoretic}]
\label{infoTheoreticLem}
Let \Atwo\ hold. Let $\X',\X{}^\tauu$ be matrices formed with disjoint subsets of the columns in $\X$. Let $\O',\O_\tauu$ indicate subsets of the observed entries in $\X'$ and $\X{}^\tauu$ with at least $\r+1$ samples per column. Suppose there are only finitely many rank-$\r$ matrices that agree with $\X'_{\O'}$, and that $\O_\tauu \in \{0,1\}{}^{\d \times (\d-\r+1)}$ satisfies condition $(\dagger)$ in Theorem \ref{mainThm}. If there is a rank-$\r$ matrix that agrees with $[\X'_{\O'} \ \X{}^\tauu_{\O_\tauu}]$, then such matrix is unique, and all entries in $[\X'_{\O'} \ \X{}^\tauu_{\O_\tauu}]$ correspond to the same $\X{}^\k$.
\end{myLemma}

The main insight behind Lemma \ref{infoTheoreticLem} is that the observed entries in $\X'_{\O'}$ impose restrictions on the rank-$\r$ matrices that may agree with the observations. The restrictions produced by $\X'_{\O'}$ may be enough to narrow the possible solutions to a finite number of options.  However, some of these restrictions may come from $\X{}^1$, others from $\X{}^2$, and so on. In such case, it is possible that the combined restrictions are compatible, leading to false rank-$\r$ matrices that agree with $\X'_{\O'}$. Incorporating $\X{}^\tauu_{\O_\tauu}$ adds more restrictions. The sampling pattern in $\O_\tauu$ guarantees that the new restrictions will add enough redundancy, such that if the restrictions do not come from the same $\X{}^\k$, they will be inconsistent, implying that no rank-$\r$ matrix can possibly agree with $[\X'_{\O'} \ \X{}^\tauu_{\O_\tauu}]$. Intuitively, $\X{}^\tauu_{\O_\tauu}$ works as a {\em checksum} matrix.

Lemma \ref{infoTheoreticLem} requires that $\X'_{\O'}$ is finitely completable. Theorem 1 and Lemma 1 in \cite{LRMCpimentel} give conditions on $\O'$ to guarantee that this is the case. We combine these results in the following Lemma, with some adaptations to our context.

\begin{myLemma}[Theorem 1 and Lemma 1 in \cite{LRMCpimentel}]
\label{finiteLem}
Let \Atwo\ hold. Suppose $\O'$ can be partitioned into $\r$ matrices $\{\O_\tauu\}_{\tauu=1}{}^\r$, each of size $\d \times (\d-\r+1)$, such that condition $(\dagger)$ in Theorem \ref{mainThm} holds for every $\tauu$. Then there are at most finitely many rank-$\r$ matrices that agree with $\X'_{\O'}$.
\end{myLemma}

To summarize: Lemma \ref{finiteLem} gives us conditions to guarantee that there are only finitely many rank-$\r$ matrices that agree with a subset of entries. If these conditions are met, Lemma \ref{infoTheoreticLem} provides further conditions to guarantee that there is only one such rank-$\r$ matrix, and that all observations come from the same $\X{}^\k$. Theorem \ref{mainThm} simply requires that each $\O{}^\k$ satisfies the conditions of Lemmas \ref{infoTheoreticLem} and \ref{finiteLem}. This way, we can just exhaustively search for all combinations of samplings that satisfy these conditions, knowing by assumption that we will eventually find $\O{}^1,\dots,\O{}^\Kk$. Then Lemmas \ref{infoTheoreticLem} and \ref{finiteLem} guarantee that we will be able to recover $\X{}^1,\dots,\X{}^\Kk$, and that we will find nothing else, i.e., no false mixtures.

\begin{proof}[\textbf{Proof of Theorem \ref{mainThm}}]
We will exhaustively search all combinations of samplings $\btilde{\O}$ with $(\r+1)(\d-\r+1)$ columns of $\O$ and $\r+1$ non-zero entries per column. For each such $\btilde{\O}$ we will verify whether it can be partitioned into matrices $\{\O_\tauu\}_{\tauu=1}{}^{\r+1}$ satisfying $(\dagger)$. If so, we will verify whether there is a rank-$\r$ matrix that agrees with $\btilde{\X}_{\btilde{\O}}$. In this case, Lemma \ref{finiteLem} implies that $\btilde{\X}_{\btilde{\O}}$ is finitely completable (because $\{\O_\tauu\}_{\tauu=1}{}^{\r}$ satisfy $(\dagger)$). Furthermore, since $\O{}^{\r+1}$ also satisfies $(\dagger)$, Lemma \ref{infoTheoreticLem} implies that $\btilde{\X}_{\btilde{\O}}$ is uniquely completable, and that all its entries correspond to the same $\X{}^\k$. It follows that $\X{}^\k$ is the only rank-$\r$ matrix that agrees with $\btilde{\X}_{\btilde{\O}}$.

By assumption, each $\O{}^\k$ can be partitioned into matrices $\{\O_\tauu\}_{\tauu=1}{}^{\r+1}$ satisfying $(\dagger)$. Hence the output of the procedure above will partition $\X_\O$ into $\X_{\O{}^1}, \dots, \X_{\O{}^\Kk}$. By \Aone\ each column in $\X_{\O{}^\k}$ has either $0$ or $\r+1$ observations, so by Lemmas \ref{infoTheoreticLem} and \ref{finiteLem} we can recover all columns of $\X{}^\k$ that have observations in $\X_\O$ using \LRMC\ techniques \cite{LRMCpimentel}.
\end{proof}

We now proceed to prove Theorem \ref{probabilityThm}, which states that if an entry of $\X{}^\k$ is observed with probability $\p=\Ord(\frac{1}{\d}\max\{\r,\log\d\})$, then with high probability $\O{}^\k$ will satisfy the combinatorial conditions of Theorem \ref{mainThm}, guaranteeing that $\X{}^\k$ is identifiable. To this end, we will use the following lemma, stating that if $\X{}^\k$ is observed on enough entries per column, then it will satisfy the combinatorial conditions of Theorem \ref{probabilityThm}.

\begin{myLemma}
\label{translationLem}
Suppose $\r \leq \frac{\d}{6}$.  Let $\epsilon >0$ be given.  Suppose that $\X{}^\k$ has at least $(\r+1)(\d-\r+1)$ columns, each observed on at least \phantomsection\label{mDef}$\m$ locations, distributed uniformly at random, and independently across columns, with
\begin{align}
\label{mEq}
\textstyle \m \ \geq \ \max \left\{ 2\r, \ 12\left( \log(\frac{\d}{\epsilon})+1\right) \right\}.
\end{align}
Then with probability at least $1-(\r+1)\epsilon$, $\O{}^\k$ satisfies the sampling conditions of Theorem \ref{mainThm}.
\end{myLemma}

Fortunately, we can prove Lemma \ref{translationLem} using Lemma 9 in \cite{LRMCpimentel}, which we restate here with some adaptations as follows.

\begin{myLemma}
\label{probabilityLem}[Lemma 9 in \cite{LRMCpimentel}]
Let the sampling assumptions of Lemma \ref{translationLem} hold.  Let $\O_{\tauu-\j}$ be a matrix formed with $\d-\r$ columns of $\O{}^\k$.  Then with probability at least $1-\frac{\epsilon}{\d}$, every matrix $\O'$ formed with a subset of the columns in $\O_{\tauu-\j}$ (including $\O_{\tauu-\j}$) has at least $\r$ fewer columns than non-zero rows.
\end{myLemma}

With Lemma \ref{probabilityLem}, the proof of Lemma \ref{translationLem} follows by two union bounds.

\begin{proof}[\textbf{Proof of Lemma \ref{translationLem}}]
Randomly select $\r+1$ disjoint matrices $\{\O_\tauu\}_{\tauu=1}^{\r+1}$ from $\O{}^\k$, each with $\d-\r+1$ columns. Let $\O_{\tauu-\j}$ denote the matrix formed with all but the $\j{}^{\rm th}$ column of $\O_\tauu$. Using a union bound and Lemma \ref{probabilityLem}, we can bound the probability that $\O_\tauu$ fails to satisfy condition ($\dagger$) by $\sum_{\j=1}^{\d-\r+1} \frac{\epsilon}{\d} \ \leq \ \sum_{\j=1}^{\d} \frac{\epsilon}{\d} \ < \ \epsilon$. Using an additional union bound, we can bound the probability that {\em some} $\O_\tauu$ fails to satisfy condition ($\dagger$) by $(\r+1)\epsilon$, as desired.
\end{proof}

All that remains is to show that if an entry of $\X{}^\k$ is observed with probability $\p$ as in Theorem \ref{probabilityThm}, then $\X{}^\k$ will be observed on enough entries per column. We show this using a simple Chernoff bound.

\begin{proof}[\textbf{Proof of Theorem \ref{probabilityThm}}]
Let $m$ be the number of observations in a column of $\X{}^\k$. Since an entry of $\X{}^\k$ is observed with probability $\p$, then $\E[m]=\d\p$, so using the multiplicative form of the Chernoff bound with $\beta=\nicefrac{1}{2}$ we get:
\begin{align*}
\Pr\left( m \leq \frac{1}{2}\d\p \right)
\ = \ \Pr\Big( m \leq (1-\beta)\E[m] \Big)
\ \leq \ e^{-\frac{\beta^2}{2}\E[m]}
\ = \ e^{-\frac{1}{8}\d\p} \leq \frac{\epsilon}{\d},
\end{align*}
where the last inequality follows because $\p \geq \frac{8}{\d}\log\frac{\d}{\epsilon}$ by assumption. This shows that with probability $\geq 1-\frac{\epsilon}{\d}$, a column in $\X{}^\k$ will have at least $\frac{\d\p}{2}=\m$ observations, with $\m$ as in \eqref{mEq}. Using a union bound on $(\r+1)\d$ columns, we conclude that with probability $\geq 1-(\r+1)\epsilon$, at least $(\r+1)\d$ columns of $\X{}^\k$ will have $\m$ or more observations, distributed uniformly at random, as required by Lemma \ref{translationLem}, which in turn implies that $\O{}^\k$ will satisfy the conditions of Theorem \ref{mainThm} with probability $\geq 1-2(\r+1)\epsilon$, as claimed.
\end{proof}

To guarantee that each $\X{}^\k$ is observed with probability $\p$, we can simply sample uniformly among $\X{}^1,\dots,\X{}^\Kk$ with probability $\Kk\p$, and hence we conclude that the sample complexity of \MMC\ is $\Ord(\frac{\Kk}{\d}\max\{\r,\log\d\})$, as claimed.

\begin{myRemark}
Notice that we cannot apply Lemma \ref{translationLem} directly instead of Theorem \ref{probabilityThm}, because if we sample $\m$ entries selected uniformly at random from each column of $\X{}^\k$, there could be {\em collisions} between multiple matrices in the mixture, which we do not allow, because that would imply observing two values for the same entry in $\X_\O$.
\end{myRemark}

\section{More about our Assumptions}
\label{assumptionsSec}
Essentially, \Atwo\ requires that $\X$ is a generic mixture of low-rank matrices. There are several equivalent ways to interpret \Aone. For instance, \Atwo\ requires that the columns in $\X{}^\k$ are drawn independently according to an absolutely continuous distribution with respect to the Lebesgue measure on an $\r$-dimensional subspace in general position. Alternatively, recall that every rank-$\r$ matrix $\X{}^\k \in \R{}^{\d \times \n}$ can be expressed as $\U{}^\k \bTheta{}^\k$, where $\U{}^\k \in \R{}^{\d \times \r}$ and $\bTheta \in \R{}^{\r \times \n}$. \Atwo\ equivalently requires that the entries in $\U{}^\k$ and $\bTheta{}^\k$ are drawn independently according to an absolutely continuous distribution with respect to the Lebesgue measure on $\R$.

\Atwo\ discards pathological cases, like matrices with identical columns or exact-zero entries, which appear with zero-probability under \Atwo. For instance, backgrounds in natural images can be highly structured but are not perfectly constant, as there is always some degree of natural variation that is reasonably modeled by an absolutely continuous (but possibly highly inhomogeneous) distribution. For example, the sky in a natural image might be strongly biased towards blue values, but each sky pixel will have at least small variations that will make the sky not perfectly constant blue. So while these are structured images, these variations make them generic enough so that our theoretical results are applicable.

Furthermore, since absolutely continuous distributions may be strongly inhomogeneous, they can be used to represent highly coherent matrices (that is, matrices whose underlying subspace is highly aligned with the canonical axes).  Typical completion theory\ \cite{candes-recht, alm, candes-tao, svt, grouse, keshavan10, recht, fpc, almNIPS, lmafit, altLRMC, lmafit2, coherent, iterative, incoherent, balzano, HRMC, ssp14, gssc, yang} cannot handle some of the highly coherent cases that our new theory covers.

However, we point out that \Atwo\ does not imply coherence nor vice-versa.  For example, coherence assumptions indeed allow some identical columns, or exact-zero entries.  However, they rule-out cases that our theory allows.  For example, consider a case where a few rows of $\U{}^\k$ are drawn i.i.d.~$\mathscr{N}(0,\sigma_1{}^2)$ and many rows of $\U{}^\k$ are drawn i.i.d.~$\mathscr{N}(0,\sigma_2{}^2)$, with $\sigma_1 \gg \sigma_2$.  This is a good model for some microscopy and astronomical applications that have a few high-intensity pixels, and many low-intensity pixels.  Such $\U{}^\k$ would yield a highly coherent matrix, which typical theory and algorithms cannot handle, while ours can. To sum up, our assumptions are different, not stronger nor weaker than the usual coherence assumptions \cite{candes-recht, HRMC, alm, candes-tao, svt, grouse, keshavan10, recht, fpc, almNIPS, lmafit, altLRMC, lmafit2, coherent, iterative, incoherent, balzano, ssp14, yang, gssc}, and we believe they are also more reasonable in many practical applications.

\section{Fine Tuning \AMMC}
\label{algorithmApx}
Section \ref{algorithmSec} presents our alternating algorithm for \MMC, summarized in Algorithm \ref{ammcAlg}. Like other mixture problems, \MMC\ is highly non-convex, and can be quite challenging in practice. In fact, to date, there exist no provable practical algorithms for even the simplest mixture problems. Arguably the most common approach is to use alternating EM-type algorithms \cite{EM, bishop, altLRMC, balzano, ssp14, gssc, constantine1}, which can only be guaranteed to converge to a local optimum, but perform well in practice. Like these algorithms, \AMMC\ also suffers from local minima. Consequently, its performance depends on initialization. In similar classification problems, it is usually convenient to initialize {\em centers} as {\em far} as possible. In our case, the centers are the subspaces containing the columns of the matrices in the mixture. Following these ideas, we initialize \AMMC\ with random subspaces as orthogonal as possible.

In addition to initialization, \AMMC\ can be further tailored to specific settings (e.g., {\bf noise}) by making small adaptations. For example, suppose instead of $\X_\O$ we observe
\begin{align*}
\X_\O \ + \ \Z_\O,
\end{align*}
where \phantomsection\label{ZDef}$\Z$ represents a noise matrix with zero-mean and variance $\sigma{}^2$.  Then, in step $\Scale[.7]{\bs{\gray{4}}}$ of \AMMC\ we can keep erasing entries of $\o$ until all the entries in $\x_{\ups{}^\k}$ are within $\sigma{}^2$ from $\bhat{\U}{}^\k$. Alternatively, one can keep in $\ups{}^\k$ only the $m$ entries of $\o$ indicating the entries of $\x_\o$ that are {\em most} likely to correspond to $\X{}^\k$, where $m$ is a tuning parameter.

Similarly, when clustering in step $\Scale[.7]{\bs{\gray{7}}}$, we can keep in $\bhat{\O}{}^\k$ only the entries of $\X_\O$ that are within $\sigma{}^2$ from $\bhat{\X}{}^\k$. Alternatively, we can keep in each $\bhat{\O}{}^\k$ only the $M$ entries corresponding to the entries of $\X_\O$ that are {\em most} likely to correspond to $\bhat{\X}{}^\k$, where $M$ is a tuning parameter that works as proxy of the noise. At the end of the procedure, the entries that not assigned to any $\bhat{\O}{}^\k$ can be considered outliers, thus providing a robust version of \MMC. In fact, this is precisely the approach that we use in our background segmentation experiments in section \ref{experimentsSec}.

Finally, if there is some side information about $\X{}^\k$, it may be beneficial to use a particular \LRMC\ algorithm in step $\Scale[.7]{\gray{8}}$ of \AMMC. For example, a two-phase sampling procedure \cite{coherent} may be better if $\X{}^\k$ is coherent. On the other hand, the inexact augmented lagrange multiplier method for \LRMC\ \cite{alm,almNIPS} is faster. Iterative hard singular value thresholding \cite{iterative} is easily implemented and often has similar performance as others \cite{LRMCpimentel}. Soft singular value thresholding \cite{svt} is better understood and has stronger theoretical guarantees. There are many other methods for \LRMC, like OptSpace \cite{keshavan10}, GROUSE \cite{grouse}, FPCA \cite{fpc}, alternating minimization \cite{altLRMC}, and LMaFit \cite{lmafit,lmafit2}, to name a few. Depending on $\X{}^\k$, it may be better to use one \LRMC\ method or an other in step $\Scale[.7]{\gray{8}}$ of \AMMC.


\end{document}